\documentclass[10pt,twocolumn,letterpaper]{article}

\usepackage{tikz}
\usepackage{pgfplots}
\usepackage[pagenumbers]{cvpr}              % To produce the preprint version
\usepackage[pdftex,pagebackref]{hyperref}
\usepackage{xcolor}
\usepackage{graphicx}
\usepackage{subcaption}
\usepackage{multirow}
\pgfplotsset{width=10cm,compat=1.9}
\usepackage{amsmath}
\usepackage{microtype}
\usepackage[accsupp]{axessibility}

\makeatletter
\newcommand{\thickhline}{%
    \noalign {\ifnum 0=`}\fi \hrule height 1pt
    \futurelet \reserved@a \@xhline
}
\makeatother

\newcommand{\ifcommentsenabled}[1]{}

\definecolor{todo_color}{rgb}{1.0,0,0.0}
\definecolor{help_color}{rgb}{0.7,0.6,0.2}

\definecolor{mathias_color}{rgb}{.6,.4,.05}
\definecolor{edited_color}{rgb}{.5,.7,.1}
\definecolor{chengcheng_color}{rgb}{0.35,0,0}
\definecolor{chris_color}{rgb}{0,0.35,0}
\definecolor{cem_color}{rgb}{0,0,0.85}
\definecolor{markus_color}{rgb}{0,0.35,0.35}
\definecolor{rob_color}{rgb}{0.35,0.35,0}

\def\cvprPaperID{1417} % *** Enter the CVPR Paper ID here
\def\confName{CVPR}
\def\confYear{2022}

\DeclareRobustCommand{\IEEEauthorrefmark}[1]{\smash{\textsuperscript{\footnotesize #1}}}

\title{DeltaCNN: End-to-End CNN Inference of Sparse Frame Differences in Videos} 

\author{
Mathias Parger\IEEEauthorrefmark{1}~~~~Chengcheng Tang\IEEEauthorrefmark{2}~~~~Christopher D. Twigg\IEEEauthorrefmark{2}~~~~Cem Keskin\IEEEauthorrefmark{2}\\ %
Robert Wang\IEEEauthorrefmark{2}~~~~~~Markus Steinberger\IEEEauthorrefmark{1}\\
\IEEEauthorrefmark{1}Graz University of Technology, \IEEEauthorrefmark{2}Meta Reality Labs\\
{\tt\small \IEEEauthorrefmark{1}\{mathias.parger, steinberger\}@icg.tugraz.at}\\
{\tt\small \IEEEauthorrefmark{2}\{chengcheng.tang, cdtwigg, cemkeskin, rywang\}@fb.com}
}

\date{June 2022}

\newcommand{\ours}{DeltaCNN\xspace}
\newcommand{\maxspeedup}{7x\xspace}

\begin{document}

\maketitle

\begin{abstract}
Convolutional neural network inference on video data requires powerful hardware for real-time processing. 
Given the inherent coherence across consecutive frames, large parts of a video typically change little.
%and thus do not need to be processed again.
By skipping identical image regions and truncating insignificant pixel updates, computational redundancy can in theory be reduced significantly. %to increase overall performance.
However, these theoretical savings have been difficult to translate into practice, as sparse updates hamper computational consistency and memory access coherence; which are key for efficiency on real hardware.
With \ours, we present a sparse convolutional neural network framework that enables sparse frame-by-frame updates to accelerate video inference in practice.
We provide sparse implementations for all typical CNN layers %(convolution, activation, pooling, etc.) 
and propagate sparse feature updates end-to-end -- without accumulating errors over time.
%Together with truncation of insignificant pixel updates,
\ours is applicable to all convolutional neural networks without retraining.
To the best of our knowledge, we are the first to significantly outperform the dense reference, cuDNN, in practical settings, achieving speedups of up to \maxspeedup with only marginal differences in accuracy.
Our CUDA kernels and PyTorch extensions can be found at \url{https://github.com/facebookresearch/DeltaCNN}.
%thus supports processing of most CNN architectures. %minimizes overhead and maximizes speedup.
%We are the first to propagate sparse feature updates end-to-end -- without accumulating errors over time -- significantly reducing overhead compared to previous work.
% Contrary to previous work, we do not forward dense feature maps between layers, but sparse feature updates instead - without accumulating errors over time.
%The code for CUDA kernels, and PyTorch extensions can be found at (for review: see supplemental).
% NN inference on videos contains much self-similarity across consecutive frames, and reprocessing same values multiple times is computationally expensive. Previous methods, such as Recurrent residual modules (RRM), have demonstrated that exploiting the similarity can reduce computation by avoiding redundant computation. However, those previous approaches did not consider hardware limitations and constraints, particularly regarding memory bandwidth for inference time, making them unpractical. We propose \ours, end-to-end sparse networks optimized for GPUs and show how we can leverage temporal similarity of videos to accelerate inference with moderate and controllable accuracy trade-offs on realistic settings. All the code used for training and evaluation could be found at (for submission: supplementary materials).
\end{abstract}

\section{Introduction}
Convolutional neural networks (CNN) are the state-of-the-art method for many image understanding tasks such as object detection, segmentation and pose estimation. 
Compared to multi-layer perceptrons, they require fewer parameters by spatially sharing parameters and perform better on image understanding tasks.
However, to address the increasing complexity of datasets and tasks, CNNs have grown to hundreds of convolutional layers requiring tens of billions of floating point operations (FLOPs).

In the last few years, researchers have found many ways to lower the cost of convolutional layers: Depth-wise separable convolutions \cite{Sifre2014}, optimizing the ratio between pixels, channels and layer count \cite{Tan2019}, quantization \cite{Moons2018, Hubara2018, Lin2016}, pruning \cite{Han2015,Le1990} and specialized hardware \cite{Han2016, Chen2015, Chen2016}, to name a few.
While these methods achieve a significant improvement in general purpose inference, there still is strong interest to further reduce the computational cost of CNNs, particularly for real-time applications on mobile devices.

Recently, researchers started to exploit the temporal similarity commonly seen in surveillance cameras, license plate recognition cameras or webcams \cite{Zhu2018,Fan2020,Nie2019,Jain2019,Li2018,Zhu2017, Pan2018,Cavigelli2020,Shelhamer2016,Habibian_2021_CVPR,Alwis2021}.
These applications often use CNNs on video input from fixed cameras, where high frame-to-frame similarity offers an orthogonal direction to reduce computational complexity.
State-of-the art frameworks for CNNs process each frame individually and therefore are not able to exploit frame-to-frame similarity.
By reusing the results from previous frames in unchanged regions, the computational cost can theoretically be reduced greatly \cite{Pan2018, Cavigelli2020, Habibian_2021_CVPR, Alwis2021} without reduction in accuracy.
%leading to power savings or higher frame rates
% \chengcheng{Maybe we can shorten or postpone the following discussions in this paragraph, or dropping them entirely?}
% Due to the nature of convolutions, updates are spread across the spatial domain (\emph{dilation}) to its neighbors with every convolutional layer.
% After a few layers, the updates will spread over the entire spatial dimension and cause dense updates for the larger part of the neural network.
% To circumvent this effect, small and insignificant updates can be truncated to retain a high sparsity throughout all layers of the neural network with only marginal differences in the final output.
% Recent publications showed that even very small thresholds can greatly increase the sparsity without negatively impacting the prediction accuracy \cite{Pan2018, Habibian_2021_CVPR}.
Furthermore, small and insignificant updates can be truncated to retain a high level of sparsity in activation throughout all layers of the CNN with only marginal differences in the final output \cite{Pan2018, Habibian_2021_CVPR}.
%Recent publications showed that even small thresholds can greatly increase the sparsity without impact on prediction accuracy 

While researchers have shown that truncating small changes increases the sparsity and thereby reduces FLOPs theoretically, leveraging data sparsity efficiently to speed up inference with actual hardware remains an unsolved challenge.
Since parallel SIMD devices, like GPUs, are typically used for CNN inference due to the advantage 
in operations per watt and memory bandwidth, it is necessary to evaluate the real-word speedup of sparse neural networks on such devices.
%of floating point operations per second and relatively high memory bandwidth compared to CPUs, it is necessary to evaluate real-word speedup of sparse neural networks with GPUs.
Specialized inference hardware as well as GPUs are less efficient for excessive conditional statements than CPUs and suffer under less structured memory access, both conditions that naturally arise when processing sparse activations in neural networks.
Thus, previous research results on sparse activation in CNNs do not translate to high speedup numbers in practice.
%At the same time, GPUs also have hardware constraints, particularly, 
%their wider-than-CPU memory bandwidths are not unbounded and 
%excessive conditional statements hinder their efficiency due to much simpler core architectures.
% \todo{emphasize that not all types of sparsity are easy to utilize on GPU to speedup inference}
% \todo{mention that other paper only look at convolutions, but not the rest}
% \chengcheng{We probably also want to mention that the benefits of memory bandwidth is an advantage of GPU, but not unbounded, since we are still resolving the memory bandwidth limitations by proposed approach.}
% \cem{I think many statements in this section are a bit broader than needed. We are leveraging data sparsity only, while there is certainly other work that leverage weight sparsity. \eg Trevor Gale's paper "Sparse GPU Kernels for Deep Learning" also shows effectively "a sparse CNN optimized for the GPU", so we should make the distinction with that line of work early on. It's not clear to the reader if the term "sparse activation" is meant to make this distinction, so we should make that clear.}
% \mathias{I was using "input" before, but that was also confusing, because I meant per layer input, not only the input to the network. I added "sparse activation" now in the last sentence of previous paragraph to introduce the term.}

In this paper, we present the first fully sparse CNN, \emph{\ours}, working on and optimized for GPUs.
Our implementation translates potential savings of sparse activation into real speedups in practice, outperforming the  state-of-the-art \emph{cuDNN} dense inference by a factor of up to \maxspeedup.
%Our implementation is on par with the speed of state-of-the-art framework \emph{cuDNN} for dense inference (out performing it in some cases), and is able to outperform their implementation by a factor of up to \maxspeedup~with sparse inference.
The main contributions of this paper are:
\begin{itemize}
    \item We propose \ours, the first sparse CNN with sparse data access end-to-end, from the input to the output for all layers, including convolution, pooling, activations, upsampling, normalization, etc.
    \ours is applicable to all CNNs with minor adaptation without retraining.
    % Our approach can be applied in parallel with other neural network acceleration methods.
    % \item We propose the first approach of sparse CNN considering the hardware advantages and limitations, and open-source the GPU implementation of sparse CNN operators. \mathias{not so sure about this one} \chengcheng{Maybe the following}
    \item{We tackle memory bandwidth and control flow issues of sparse neural networks by a new kernel design involving masks and caches. We open source, to the best of our knowledge, the first GPU implementation of CNN operators for sparse input and output.}
    % For fully utilizing the potential of GPUs while respecting their limitations in these sparse operators, we propose mask-based cache loading mechanism and reduction of conditional statements.
    \item We show the first GPU-based demonstration of leveraging data sparsity for CNN acceleration by speeding up three networks for object detection and human pose estimation on three types of GPUs by up to \maxspeedup.
    % \todo{rephrase - not sure if that statements holds against CBInfer}
    % \item We present the first fully sparse convolutional neural network on GPUs. Our implementation uses the sparsity to speedup convolutions as well as pooling operations, activations, upsampling, normalization, etc.
    % \item We propose a new approach to truncate small updates to increase sparsity without loss in accuracy and with low computational and memory overhead. Our approach can be applied on every existing CNN without retraining.
    % \item \todo{remove this point?} We evaluate our approach on 3 different neural network architectures and 3 data sets for object detection and human pose estimation.
\end{itemize}
Our evaluations on three GPU architectures show that \ours~is efficiently implemented, matching the speed of \emph{cuDNN} when operating without sparsity.
In sparse mode, we achieve speedups of up to \maxspeedup~over \emph{cuDNN}.

\section{Related Work}
Recent work on exploiting frame-to-frame similarity in videos can coarsely be divided in two groups: optimized model architectures and exploiting feature sparsity by truncating insignificant updates.

\subsection{Efficient video CNN architectures}
Efficient CNNs aim to reduce the frequency in which the most expensive part of the network, the backbone, is processed.
Two path models use a fine-grained feature generation on key frames and a coarse-grained update path for frames in between~\cite{Fan2020,Nie2019}.
Alternatively, the fine-grained features can be adapted directly, \eg by using optical flow of the network input~\cite{Jain2019, Zhu2017}.
% \cem{There is a command eg or Eg for correct typing of e.g. Let's use that consistently.}
Our approach does not require any changes to the network architecture and automatically performs fine-grained updates where required.
% However, direct feature map updates could potentially be applied on top of our approach to increase reuse of processed features for panning cameras, as discussed in Section \ref{sec:conclusion}.

\subsection{Sparsity in videos}
\begin{figure}[pt]
\centering
\includegraphics[width=0.95\linewidth]{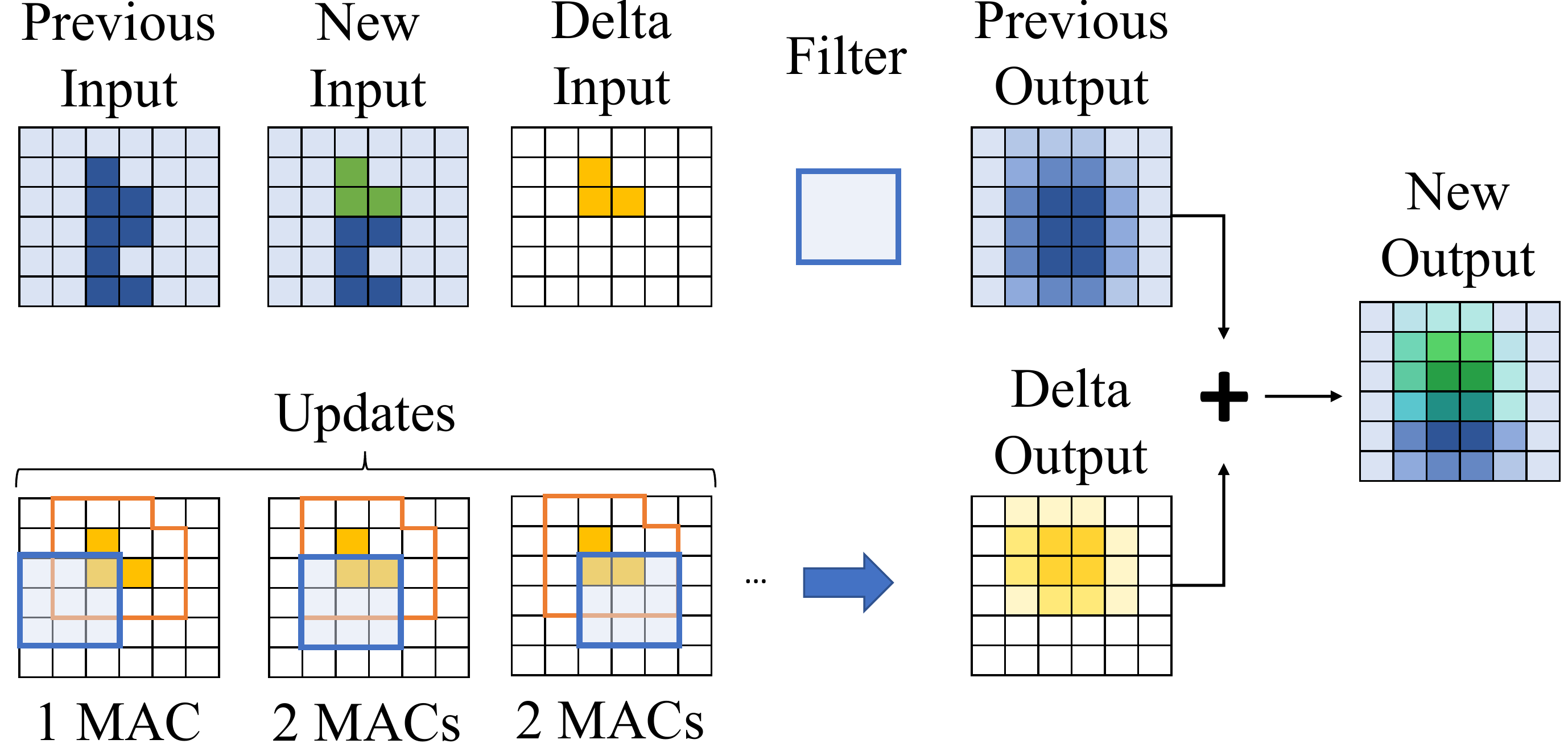}
\caption{Working principle of spatially sparse convolutions for videos. 
Computing the difference between the current and previous input, large parts of convolution input become zero (white).
%By subtracting the previous input from the input of the current frame, large parts of convolution input become zero (white).
Since zero valued inputs do not contribute to the output, these values can be skipped to reduce the number of operations.
%multiply-accumulate operations (MACs). 
%\chengcheng{We might need to think more about how to present this in relationship with the sub-tile based approach: what we illustrated here is for sparse sub-tile convolution; while we also show later that sub-tile sparsity is challenging due to conditional statements. I think one way to resolve this is to just note that there are more to consider than this simple principals here, and refer pargraphs of text?}
}
\label{fig:delta_conv_sketch}
\end{figure}

Data sparsity in CNNs can be understood as zero-valued features in the feature maps.
While activation functions like ReLU already lead to some level of feature sparsity, sparsity in videos can typically be increased greatly by using the difference between the current and the previous frame as input (see Figure \ref{fig:delta_conv_sketch}).
This way, background and static features become zero and can be skipped. %while only updated features need to be reprocessed.
This characteristic is utilized both in 2D \cite{Pan2018,Cavigelli2020,Habibian_2021_CVPR,Alwis2021} and in 3D \cite{Peng_2021_WACV} CNNs.

\textbf{Update truncation}
\emph{Recurrent Residual Module (RRM)} \cite{Pan2018} and \emph{CBinfer} \cite{Cavigelli2020} show that sparsity can be increased even further by truncating insignificant updates without significant loss of accuracy.
Contrary to them, \emph{Skip-Convolution} \cite{Habibian_2021_CVPR} does not truncate input features, but output features instead.
This can lead to higher sparsity, but requires dense updates at a regular schedule (4-8 frames).
%They use gates to decide which pixel of the feature map requires an update depending on the feature updates on the input.
%However, this approach has the disadvantage of requiring dense updates at a regular schedule (4-8 frames) as continuous updates of internal buffers is not possible with output gating. 
We use a combination of these ideas.
Like \emph{Skip-Convolution} and \emph{CBInfer}, we use a spatial (per pixel) sparsity, since structured sparsity better suits SIMD architectures than per value sparsity. 
% \cem{I can't follow the reasoning in this sentence. We "use per pixel sparsity" but "per value sparsity" is bad?}
% \mathias{on a GPU, 32 threads always execute the same instruction. In case of convolutions, this works much better when all 32 threads work on the same pixel, but different channels, as they will always have the same behaviour when handling tile borders etc. If any of the channels is sparse however, this would not make a difference (some threads would process the values, the rest idles) or at least require a huge overhead to find a set of 32 values that were updated.}
Like \emph{RRM}, we decide per \emph{input} pixel if an update is required.
%If the largest value of a pixel is larger than a threshold, all output pixels affected will be updated.
While having a single input pixel triggering an update of many output pixels may increase FLOPs compared to \emph{Skip-Convolution}, it is crucial in enabling a continuous inference in sparse mode without accumulating errors over time.
% In contrast, \emph{Skip-Convolution} requires dense updates at a regular schedule (4-8 frames) and the sparse updates in between are always applied on the latest dense reference frame, as continuous updates of buffers are not possible with output pixel gating.

% This way, we achieve a good sparsity/accuracy loss trade-off without the need of running additional kernels on the GPU, as this can easily be combined with other operations like activation functions.

\textbf{Caching the previous state}
\emph{RRM}, \emph{CBInfer} and \emph{Skip-Convolution} cache input and output feature maps from previous frames at every convolutional layer to process the differences, increase sparsity and then accumulate the outputs together with dense outputs from previous frames. 
% \todo{cbinfer also accelerates pooling operations - I mention that in Section Sparse Inference}
All operations between convolutions are processed densely.
While this strategy reduces FLOPs, it increases memory transfer. %using accumulated values.
To reduce the memory overhead, \cite{Alwis2021} proposed to only store input and output buffers on key convolutional layers and only use frame differences in between.
%the operations in between directly on the frame-delta.
This approach fails for non-linear layers like pooling or activation functions and may lead to significant errors (see Section~\ref{sec:ours}). 
%However, this comes with a great decrease in accuracy, as many operations like pooling and activation layer are nonlinear, and return incorrect results when applied on frame-differences directly (see Section \ref{sec:ours}).
\ours~performs all operations sparsely, by comparing only the input features (camera image) against the complete previous input, propagating the sparse feature updates throughout all layers.
The dense results are only accumulated at the final layer. 
This way, we also accelerate non-convolutional layers like pooling, upsampling and activations.
Furthermore, we avoid switching between sparse and dense computations and only need to cache accumulated values for nonlinear layers, %we also reduce the overhead of moving from dense accumulated values to sparse updates and back for every convolutional layer.
%We only need to cache accumulated values for nonlinear layers,
reducing the number of caches compared to~\cite{Pan2018,Cavigelli2020,Habibian_2021_CVPR}, without loss in accuracy.

\begin{figure*}[t]
\centering
\includegraphics[width=0.9\textwidth]{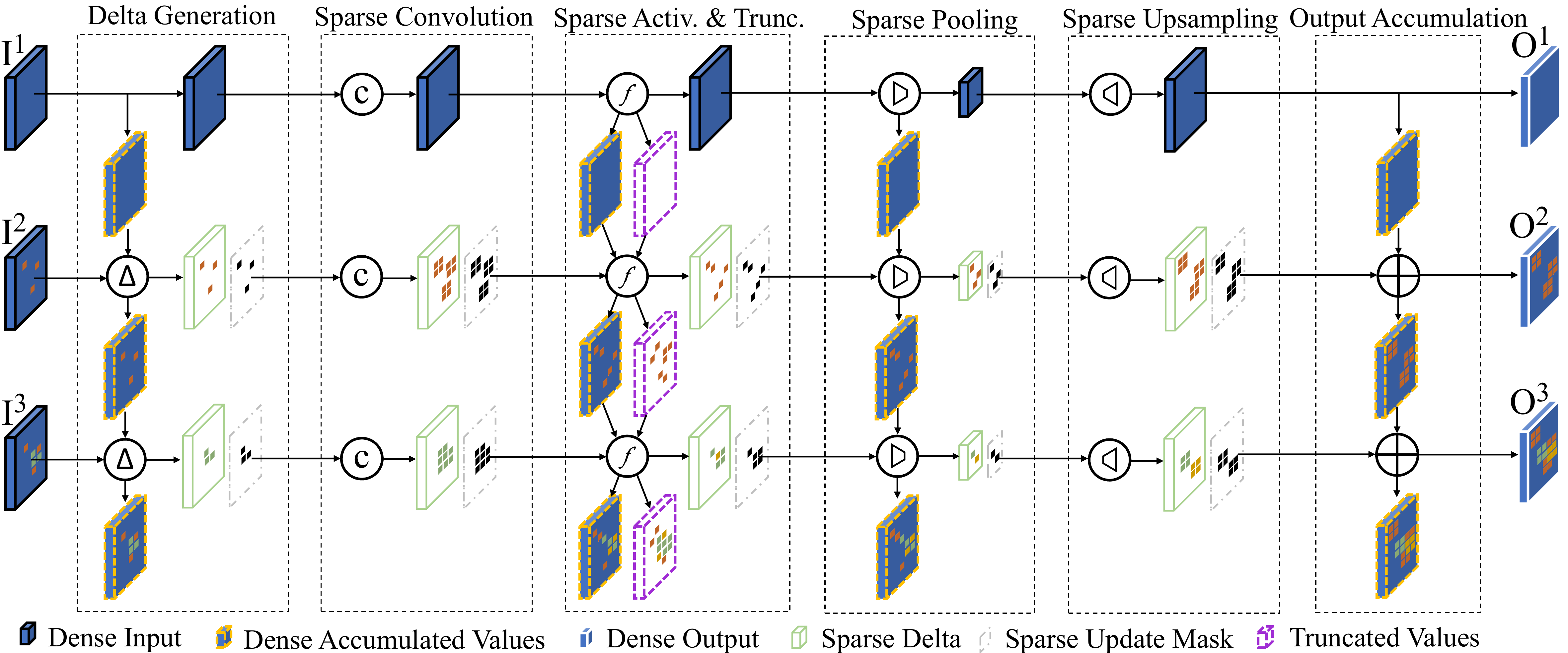}
\caption{Illustration of inference over three frames using \ours~for a toy network that consists of convolution, activation, pooling, and upsampling layers. The first frame $I^{1}$ is processed densely and is used to initialize the buffers for \emph{Dense Accumulated Values}. In subsequent frames, \emph{Delta Generation} subtracts the previous input from the current to generate an \emph{Update Mask} and a \emph{Sparse Delta} feature map containing only significant pixel updates. After \emph{Sparse Convolution}, causing the update mask to dilate, \emph{Sparse Activation \& Truncation} truncates small values to increase sparsity. After the final layer, the \emph{Sparse Delta} output is accumulated onto the previous output buffer to generate a \emph{Dense Output} $O^{i}$ for frame $I^{i}$. 
% It is worth noting that end-to-end sparsity of input and output is kept when adding additional layers.
}
% , which also do not need dense access to caches.}
\label{fig:delta_cnn}
\end{figure*}

\subsection{CNN kernels for sparse data}
% \cem{Sparse CNN "kernel" sounds a lot like a sparse filter, as in sparse weights, like in Trevor Gale's paper. "Data sparsity in CNNs" or something like that is less ambiguous.}
% \mathias{What about "CNN kernels for sparse data"?}
% \todo{restructure: e.g. paragraph for memory, one for sparsity, etc.}
% \todo{rewrite and shorten once story is complete}
% \todo{mention some things that later explain in 3.2 - memory considerations, tiles, etc.}
% \mathias{should we remove details about SBNet and SSC?}
While \emph{RRM} and \emph{Skip-Convolution} showed that sparsity in videos can be used to significantly lower FLOPs, they could not show that their approach translates to wall-clock time improvements compared to state-of-the-art dense CNN frameworks. 
Existing sparse convolution implementations like \emph{Sparse Blocks Network (SBNet)} \cite{Ren2018} and \emph{Submanifold Sparse Convolutional Network (SSC)} \cite{Graham2018} utilize data sparsity to accelerate inference, but are not designed for video inference.
Both of these methods assume that the input data is inherently sparse, like handwriting or object boundaries in 3D volumes for SSC, or semantic segmentation masks for SBNet.
They do not utilize any mechanism like a cache for storing the dense state of previous frames, and can therefore not be applied on videos.
% They do not utilize any mechanism like a cache, which is needed to process the full frame and not just the changes.
% \emph{SSC} implements a native sparse convolution, processing only values that were updated.
% However, their approach is optimized for sparse data like thin lines or handwriting, and therefore does not support dilation, which is critical for many dense image understanding tasks like segmentation, object detection or pose estimation from camera images.
% \emph{SBNet} uses a user-defined mask and a feature map as input to gather updated blocks into a smaller image, performs a dense convolution and afterwards scatters the results back to their corresponding output positions.
% Combined with before mentioned sparse video approaches, \emph{SBNet} can be used to accelerate convolutions by dropping blocks without updates.
\emph{Pack and detect} \cite{Kumar2019} performs full convolutions only on key frames and continuous updates on a smaller image containing the previously detected regions of interest.
Unfortunately, the packing operations lead to a large overhead to every convolutional layer.
%With the requirement to run many additional kernels per convolution, it adds a large overhead to every convolutional layer.
\emph{CBInfer} implements general sparse convolutions and pooling operations optimized for video input.
They perform change detection, change indexing, feature gathering, convolution and feature scattering operations for each convolution.
This allows them to utilize fine grained sparsity, but comes with significant data movement overhead and requires convolution algorithms which are inferior compared to leading implementations like \emph{cuDNN}.
In contrast, our approach performs the convolution directly on the feature maps without the need for pre- or post-processing.
Together with sparse feature maps, we propagate update masks between layers -- starting computations only where necessary. %images were updated and which can be skipped.

\section{Method}
We propose \ours, an end-to-end sparse CNN framework for accelerating video inference by exploiting frame-to-frame similarity.
\ours~replaces all dense tensor operations with sparse operations using an update mask to track which pixels to process.
% We further propose a method to increase the sparsity with minimal changes in the network output.
% By replacing all dense tensor operations with sparse operations based on caches and update masks,
% By propagating an update mask together with the feature map, all layer in a CNN can operate sparsely, only processing updated values. 
\ours~increases the sparsity with minimal changes in the network output, and processes only the sparse frame updates for each layer.

\subsection{Delta value propagation}
\label{sec:ours}

% At the core of our method, we need to keep track of previous frames' intermediate results at each layer, process the difference, then add the differences of inputs and outputs back to previous results.
The core feature of \ours~is to propagate sparse frame updates through the network end-to-end (see Figure~\ref{fig:delta_cnn}).
To reuse computation of previous frames by adding update tensors, we need to support both linear (\eg convolutions) and nonlinear (\eg activations) layers in a CNN.

\textbf{Linearity of convolutions}
Convolutions ($c$) are linear operators (see Figure \ref{fig:delta_conv_sketch}), \ie
\begin{equation}
c(x+\delta x) = c(x) + c(\delta x).
\end{equation}
% \cem{Let's use the ie command rather than i.e. I fixed it everywhere.}
This allows us to use the difference between two images, called \emph{delta}, as input to the convolution.
% \emph{Delta} outputs can be fed as input to consecutive convolutions without the need to accumulate individual outputs between two convolutions.
\emph{Delta} outputs can be fed as input to consecutive convolutions without the need to accumulate delta updates over multiple frames.

\textbf{Nonlinear layers}
Most activation functions, however, are nonlinear. 
% \todo{Fixed non-linear to nonlinear for consistency. We should check everywhere.}
% and therefore have to be performed with total values. \chengcheng{Can we put a very small illustration to demonstrate this?}
For example, the ReLU activation is defined as:
\begin{equation}
f_{ReLU}(x) = max(x, 0).
\end{equation}
The nonlinearity of the activation function poses a challenge to update previous results by \emph{delta} updates.
For example, 
\begin{equation*}
f_{ReLU}(-1) + f_{ReLU}(2) \neq f_{ReLU}(1).
\end{equation*}
To solve this challenge, we keep track of accumulated inputs to nonlinear layers.
% However, this defeats the purpose of sparse convolution since it implies that a dense operator needs to be carried out to add the pre-activated values together.
% While the \emph{delta} input likely includes negative values, these values do not necessarily have to be clamped to zero, since the total value could still be positive. 
% Likewise, even a positive value could still be clamped due to the total value being negative.
% Thus, nonlinear activation functions, as well as maximum pooling layers, are performed with total values.

% \paragraph{Buffer for nonlinear layers}

For a given activation or pooling function $f$, the \emph{delta} output $\delta y$ is defined as
\begin{equation}
    \delta y = f(x+\delta x) - f(x)
\end{equation}
with $\delta x$ being the \emph{delta} input.
% and $\delta y$ being the \emph{delta} output.
% $a$ being the accumulated layer input.
The difference, $\delta y$, is then used as \emph{delta} input for subsequent layers.  
Every nonlinear layer stores its own buffer holding the previous accumulated inputs.
The buffers are initialized during the first frame performing dense inference, and kept up-to-date using the \emph{delta} of the following frames.
Previously accumulated inputs implicitly include all biases applied before them.
Thus, biases in convolutional and batch normalization layer are only applied in the first frame.

\begin{figure}[t]
\centering
\includegraphics[width=\linewidth]{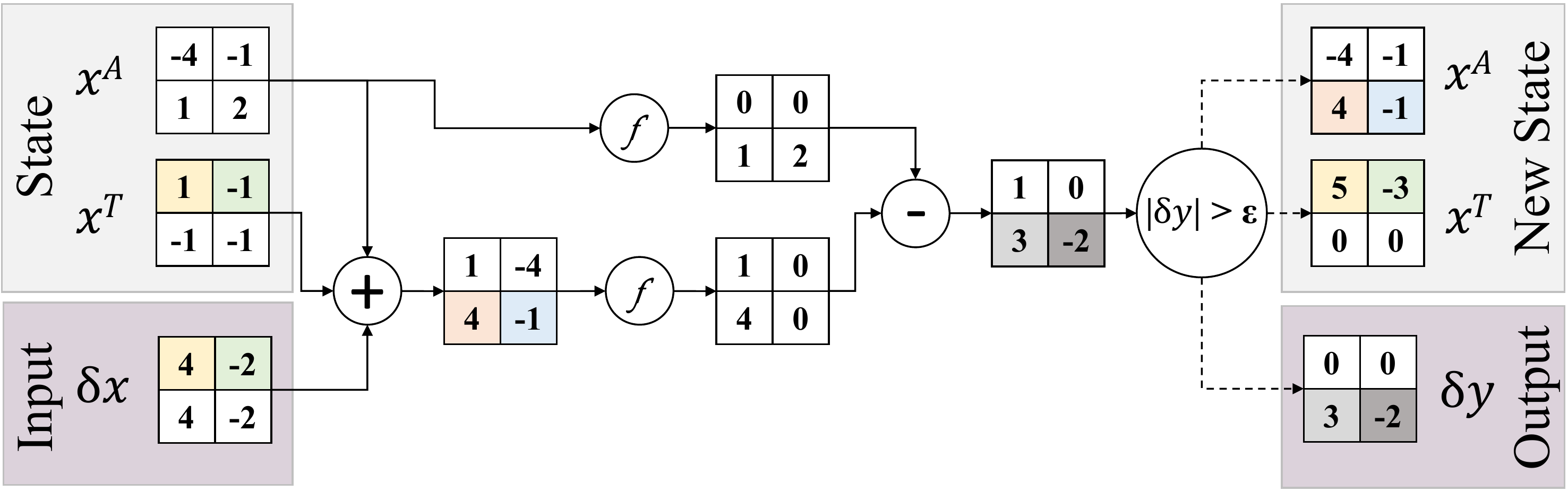}
\caption{Illustration of the activation and truncation function.
It performs activations with accumulated values, truncates the \emph{delta} $\delta x$ output, and updates the accumulated $x^A$ and truncated $x^T$ values buffers.
In this example, \emph{ReLU} is used as activation function $f$ and the truncation threshold $\epsilon$ is set to $1.5$. 
Colored tiles indicate relations between inputs, updated state and output values.
% \todo{put a and t on top and bottom and only d on the sides, because a and t are only internal state or vice versa}
% \todo{think about how we could animate this in the video!}
% \todo{in the video, we could 2 variants: 1 with and 1 without truncated values buffers}
% \todo{write State and Input vertically to save space}
}
\label{fig:activate_truncate_sketch}
\end{figure}

\textbf{Truncating small updates}
% Activation layers are very common in CNNs - nearly every convolutional layer is followed by an activation layer.
% Combined with the fact that they need to read and write the entire feature maps, we perform activation and value truncation together in one step to minimize truncation overhead.
Since activation functions follow most convolutional layers, and need to operate on every pixel in the feature map, combining activation and truncation into a single operation helps to minimize overhead.
The decision about which values can be truncated is made on a per-pixel level; we truncate a given pixel (setting all channels to 0) and mark it unchanged if $\max_k |\delta y_k| < \epsilon$. 
%If any $|\delta y_i|$ is larger than $\epsilon$, the pixel is marked as changed and the previous accumulated values buffer $x^A$ is updated using
If any $|\delta y_k|$ is larger than $\epsilon$, the pixel is marked as updated, and we use an accumulated values buffer $x^A$ to store the current value at frame $i$,
\begin{equation}
\label{eq:update_prev_buffer}
    x^A_{i} = x^A_{i-1} + \delta x.
\end{equation}
However, small truncations can add up over time and lead to a decrease in accuracy with every frame.
\Eg, when the sun is slowly rising, lighting up an outdoor scene, the frame to frame differences are too small to trigger an update, and the accumulated errors cannot be corrected after being dropped once.
To solve this issue, we introduce a second buffer $x^T$ containing the accumulated truncations since the last update.
%To solve this issue, we store two buffers for every layer that performs truncation: an accumulated values buffer $x^A$ containing the state of accumulated values when a pixel was last updated, and $x^T$ containing accumulated truncations since the last update.
The truncated values $x^T$ are used together with the delta values $\delta x$ and the accumulated values $x^A$ in activation functions:
\begin{equation}
   \delta y = f(x^A+x^T+\delta x) - f(x^A).
\end{equation}
When a pixel is truncated, $\delta x$ is added onto the truncated values buffer $x^T$.
When a pixel is marked as updated, $x^A$ is updated using
\begin{equation}
\label{eq:update_prev_buffer_truncated}
x^A_{i} = x^A_{i-1} + x^T_{i-1} + \delta x
\end{equation}
and the truncated values buffer $x^T_i$ is set to zero (see Figure \ref{fig:activate_truncate_sketch}).
Using this technique, \ours~only requires one dense initial frame, and can apply sparse updates indefinitely without accumulating errors over time. 
% \cem{We never mention what the index i is here. And we have versions of $x^T$ with and without the index, which is not clarified.}

\subsection{Design considerations for GPUs}
General purpose GPUs allow for the execution of arbitrary code on many core devices. % like on CPUs.
Yet, theoretical instruction reductions often do not translate into efficient GPU code.
%Yet, software written and optimized for CPUs can often not be directly translated into efficient GPU code.
%Compared to a CPU, GPUs have much smaller per-core caches and less sophisticated control units.
Non-consistent execution paths and incoherent memory access may lead to significant slowdowns.
%With smaller per-core caches, and much less sophisticated control units to minimize execution latency, the key to achieve good convolution performance on GPUs is to optimize memory accesses and distribute the work into small and independent units for high parallelism.
The key to high convolution performance on GPUs is to optimize memory accesses and generate locally coherent control flow.
%the work into small and independent units for high parallelism.

\textbf{Update mask}
% a sentence to say that previous work do some dense comparison?
% Loading previous cached results densely for detecting changes leads to excessive memory access.
While the changes propagated through the network are sparse, we store the delta features as dense tensors of the same shape in our GPU implementation; with zeroes for locations that can be ignored.
To take a previous layer's sparse output as input, every layer needs to know which pixels were updated.
One approach to do this would be a zero-check on all values of the delta input at the beginning of each layer as done by previous work \cite{Habibian_2021_CVPR,Pan2018,Cavigelli2020}. 

To avoid loading and checking entire inputs for updates, we propagate -- together with the delta feature map -- a spatial update mask which contains one value per pixel, indicating if it was updated or not.
For every layer, before loading any other data, we first check the update mask of all input pixels and for an entire tile (see below) decide whether to skip all memory operations and computations.
Independent of whether a tile is skipped, we write the update mask for the subsequent layer.
%performs or skips the execution of the current tile (see below) and returns a new update mask for subsequent layers.
% This way, we reduce memory bandwidth and can terminate processing a tile earlier, freeing up GPU resources which can then be used to process subsequent tiles.
% This way, we reduce memory bandwidth by loading only processed pixels and speed up the computation by avoiding operations related to zero-checking.
%This way, we reduce memory bandwidth and detect empty tiles early, freeing up GPU resources which can then be used to process subsequent tiles.
Using the update mask, we do not need to initialize unprocessed values in feature maps to zero, as they are never read, reducing memory bandwidth even further. 

\textbf{Memory considerations and tiled convolutions}
Convolutional layers typically make up the majority of processing time in a CNN.
While we provide optimized implementations for various layer types, we focus our design discussions around convolutions.
These considerations naturally translate to other layer types.
%While it is important to support all layers for efficient sparse inference,  we focus our design discussions around convolutions for simplification.

A common approach to optimize memory reuse and locality in 2D convolutions is to process the image in tiles, with each tile processed by a cooperative thread array (CTA).
This way, input features and filter parameters can be kept local and reused multiple times.
The tile size is chosen in a way to balance the trade-off between memory access and the level of parallelism.
Larger tiles reduce memory access but require more resources.
%-- the larger the tile size, the less CTAs can run simultaneously, potentially leading to low utilization of the available hardware.

Previous work on sparse convolutions focused on FLOP reductions, but hardly showed wall-clock time improvements in practice.
Utilizing sparsity on a fine grained level to avoid unnecessary multiplications requires many additional conditional jumps and may lead to scattered memory operations, easily costing more time than they save, even when the majority of FLOPs can be skipped.

% There is a trade-off between the sparsity on a tile level and sparsity at a sub-tile level.
\textbf{Per-tile sparsity vs.\ sub-tile sparsity}
Instead of fine-granular conditions, we employ sparsity on a tile level.
%The best way to utilize sparsity to one's advantage is to skip processing entire tiles.
In the case of even only one pixel of a tile being updated, the cost is nearly as high as when all pixels are updated, since all filter parameters need to be loaded and multiple output pixels need to be processed and written.
For example, consider a tile size of 5x5 output pixels, a 3x3 convolution kernel with 256 channels, requiring a 7x7 input.
%The 3x3 convolutional kernel takes 256 channels as input and produces 256 output channels.
In this example, one CTA loads 12,544 input features and 589,824 filter parameters and performs 14,745,600 multiplications per tile.
If any of the inputs are non-zero, most of the memory transfers are still required; only the number of multiplications could potentially be reduced.
%However, efficient GPU implementations require the code to be as static as possible, which requires us to simplify the control flow.

\textbf{Control flow simplification}
% In the best case, the exact set of multiplications is known at compile time, as this allows for more efficient register usage and loop unrolling.
% Input values, filters and intermediate results can be stored in registers, thus reducing the number of memory or cache transfers.
Convolution operations involve inputs, kernels, and outputs;
knowing the association of the three components and their memory location in registers at compile time makes the computation more efficient.
Excessive conditional control flows deciding whether to skip a pixel at a sub-tile level require loading variables, performing comparisons, and conditional jumps.
This increases the number of executed instructions many times, deteriorating performance.

To avoid fine granular conditional jumps, we propose a hybrid kernel, deciding between three processing modes: skip, dense and very sparse.
%For best performance for dense tiles as well as very sparse tiles, we use a hybrid kernel that selects the best suited approach for the given tile density.
% Tiles with 5 or more input pixel updates are processed densely, i.e. as if all inputs were updated, in order to be able to use the very efficient implementation without any conditional jumps.
Tiles with no active input pixels skip all loads and computations.
Tiles with five or more active input pixels (out of up to 64) are processed densely without any conditional jumps.
Tiles with one to four updated input pixels use a special, highly optimized kernel: it iterates only over a short array of updated pixels gathered from the update mask, removing the need to check the update mask of all pixels many times during multiplication.
Furthermore, in this mode we only load filter weights that are required for processing a specific tile.
This way, we can reduce memory transactions by up to 8x, \eg when there is only a single update in the top left corner and it will only affect the top left output pixel.
The advantages and disadvantages of sub-tile sparsity are further evaluated in the supplemental material.

\subsection{Truncation of insignificant updates}
\label{sec:truncation}
Convolutions dilate the update mask with every layer. 
A single pixel update in the input quickly expands to 49 pixels after three 3x3 convolutional layers.
As not all updates contribute equally to the output of the network, we truncate insignificant updates to increase sparsity and thereby speedup inference.
\ours compares the maximum norm of a pixel with the threshold $\epsilon$ to determine if the pixel update can be truncated.

Typically, the ideal $\epsilon$ will vary between networks and even layers.
We auto tune each layer's $\epsilon$ in a front-to-back manner on a small subset of the training set. 
% % While some networks can achieve high sparsity with a uniform threshold $\epsilon$, we achieved best results by auto tuning the threshold per layer in a front-to-back manner.
%With a predefined margin of error increase per truncation layer, we iteratively increase the threshold as long as the error (\eg loss), averaged over a subset of the training set, stays blow the margin.
Starting with a low $\epsilon$,  we iteratively increase the layer's  $\epsilon$ as long as the loss stays below a predefined margin of error, \ie, we allow each truncation layer to contribute equally to the output error.
Once the highest threshold below this margin is found, we freeze that layer's $\epsilon$ and continue with the next in order of execution.
Experiments show that we also need to limit the increase in accuracy when tuning for thresholds to avoid overfitting on the small subset of the training set.

\subsection{Implementation}
% \ours~provides a large collection of common CNN layers in various configurations.
% Converting the \emph{deltas} to \emph{total} values and back adds computational and memory overhead.
With the goal of not only accelerating convolutions, but the entire network, performing as many operations sparsely as possible is crucial.
Hence, \ours~provides sparse implementations for most common layers in today's CNNs: convolutions, batch normalizations, pooling layers, upsampling layers, activations, concatenations and additions.
We provide CUDA kernels (as cuDNN replacement) and PyTorch extensions that can be used as a direct replacement for the corresponding PyTorch layers, reusing the original parameters and model logic.
\section{Evaluation}
We evaluate \ours on two common image understanding tasks: human pose estimation (Human3.6M\footnote{The Human 3.6M data was received and exclusively accessed by Mathias Parger. Meta did not have access to the data as part of this research.}\cite{Ionescu2014}) and object detection (MOT16\cite{Dendorfer2021} and WildTrack\cite{Chavdarova2018}).
In both cases, we trained CNNs on video datasets using pre-trained weights from image datasets. 
After training, convolutional layers and batch normalization layers were fused where possible to improve performance both for the baselines as well as for \ours.
%For \ours, we are able to replace all operations with our sparse counterparts.
%We replace all operations performed on the feature tensors by our sparse counterparts.
% , i.e. convolution, activation, pooling, upsampling, multiplication, addition, concatenation layers and remaining batch normalization layers.
Only the first and last layer for \ours operate on dense data, converting the dense video input to sparse \emph{delta} features and converting from \emph{delta} to dense accumulated outputs, respectively.
%Additionally, we add a layer converting the dense video input to sparse \emph{delta} features before the first layer and inverse conversions from \emph{delta} to accumulated features at CNN outputs.

\begin{figure}[pt]
% inputs
% \begin{subfigure}{.33\linewidth}
%   \centering
%   \includegraphics[width=.95\linewidth]{figures/update_masks/mot16/video_02_21.png}
% \end{subfigure}%
% \begin{subfigure}{.33\linewidth}
%   \centering
%   \includegraphics[width=.95\linewidth]{figures/update_masks/mot16/mask_02_21.png}
% \end{subfigure}%
% % \begin{subfigure}{.2\linewidth}
% %   \centering
% %   \includegraphics[width=.95\linewidth]{figures/update_masks/mot16/updates_02_21.png}
% % \end{subfigure}%
% \begin{subfigure}{.33\linewidth}
%   \centering
%   \includegraphics[width=.95\linewidth]{figures/update_masks/mot16/mask_02_21_no_dil.png}
% \end{subfigure}%
% % \begin{subfigure}{.2\linewidth}
% %   \centering
% %   \includegraphics[width=.95\linewidth]{figures/update_masks/mot16/updates_02_21_no_dil.png}
% % \end{subfigure}%

% \begin{subfigure}{.33\linewidth}
%   \centering
%   \includegraphics[width=.95\linewidth]{figures/update_masks/mot16/video_04_21.png}
% \end{subfigure}%
% \begin{subfigure}{.33\linewidth}
%   \centering
%   \includegraphics[width=.95\linewidth]{figures/update_masks/mot16/mask_04_21.png}
% \end{subfigure}%
% % \begin{subfigure}{.2\linewidth}
% %   \centering
% %   \includegraphics[width=.95\linewidth]{figures/update_masks/mot16/updates_04_21.png}
% % \end{subfigure}%
% \begin{subfigure}{.33\linewidth}
%   \centering
%   \includegraphics[width=.95\linewidth]{figures/update_masks/mot16/mask_04_21_no_dil.png}
% \end{subfigure}%
% % \begin{subfigure}{.2\linewidth}
% %   \centering
% %   \includegraphics[width=.95\linewidth]{figures/update_masks/mot16/updates_04_21_no_dil.png}
% % \end{subfigure}%

\begin{subfigure}{.33\linewidth}
  \centering
  \includegraphics[width=.95\linewidth]{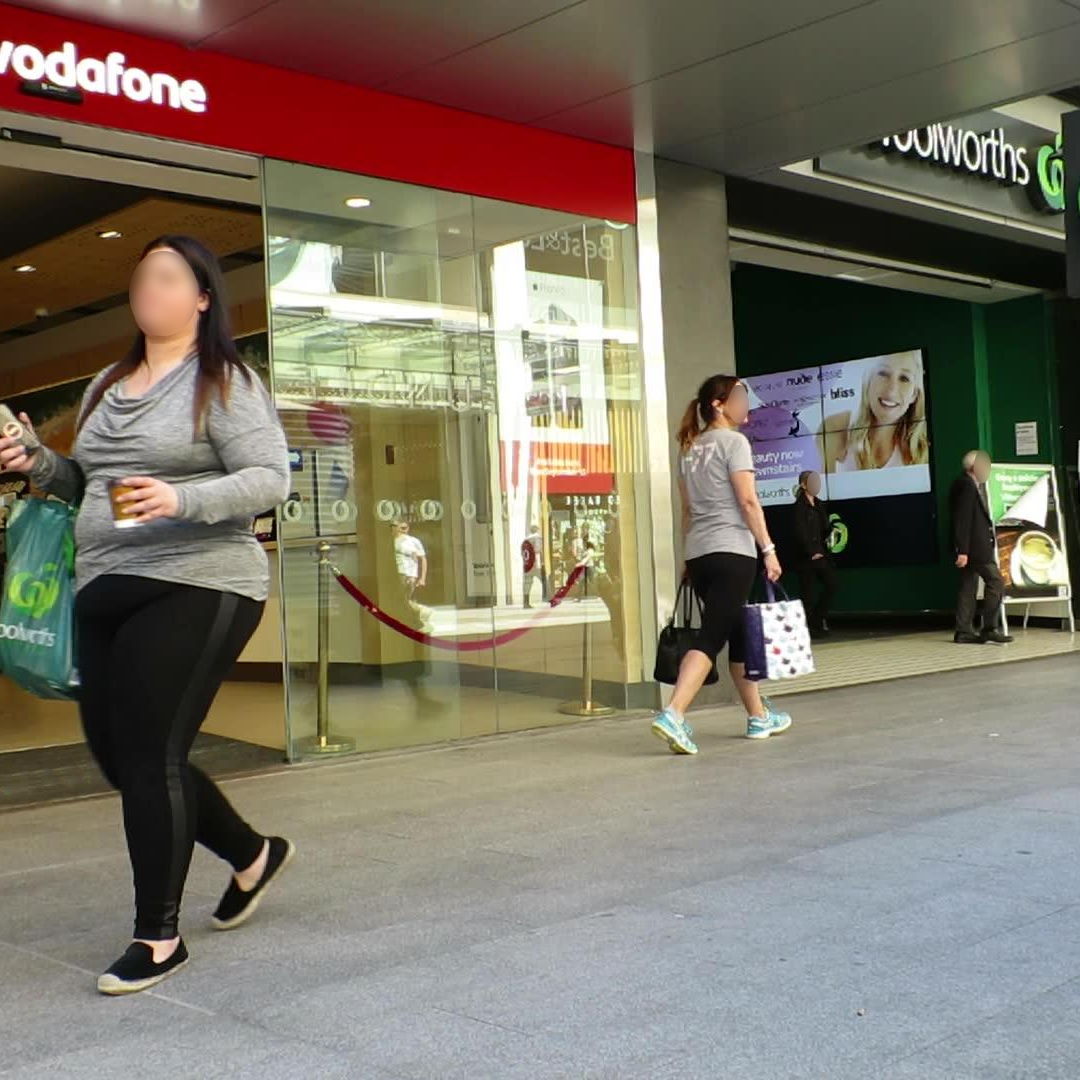}
  \caption*{Input}
\end{subfigure}%
\begin{subfigure}{.33\linewidth}
  \centering
  \includegraphics[width=.95\linewidth]{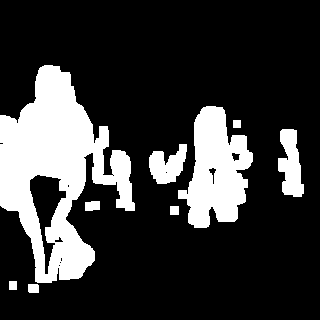}
  \caption*{(a) Mask w/ dilation}
\end{subfigure}%
% \begin{subfigure}{.2\linewidth}
%   \centering
%   \includegraphics[width=.95\linewidth]{figures/update_masks/mot16/updates_09_21.png}
%   \caption*{(a) Updates}
% \end{subfigure}%
\begin{subfigure}{.33\linewidth}
  \centering
  \includegraphics[width=.95\linewidth]{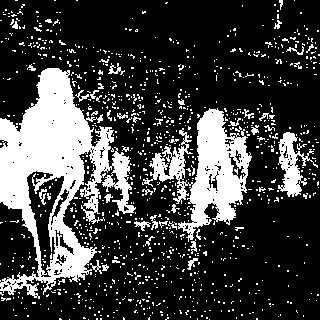}
  \caption*{(b) Mask w/o dilation}
\end{subfigure}%
% \begin{subfigure}{.2\linewidth}
%   \centering
%   \includegraphics[width=.95\linewidth]{figures/update_masks/mot16/updates_09_21_no_dil.png}
%   \caption*{(b) Updates}
% \end{subfigure}%

\caption{Input masks generated using a threshold of $\epsilon=0.5$ and 7 pixel dilation (a) and using a threshold of $\epsilon=0.15$ without dilation (b). The dilated version runs 8\% faster and achieves slightly more accurate results on the MOT16 dataset.
% The average update density per pixel shows that the dilated version requires less updates. \chengcheng{It would be worthwhile to add another image just to compare the qualitative results of detection with mot16.}
% \mathias{An image with the predicted anchors? Could do that, but I assume it will look identical.}
}
\label{fig:update_mask_mot16}
\end{figure}

%The truncation thresholds are auto-tuned front-to-back for every network and dataset separately according to Section \ref{sec:truncation}.
In all cases, we use multiple randomly selected training sequences, each consisting of 100 frames, and average the loss over all frames using  
auto-tuned $\epsilon$ thresholds (as described in Section~\ref{sec:truncation}).
%currently selected thresholds.
The maximum loss increase over all layers in total is set to 3\%, with each layer only allowed to increase the loss by a fraction of this value.
For the first truncation threshold, \ie, the input video normalized on ImageNet \cite{Deng2010} color range, we use an increased threshold to suppress the background noise, but make sure to stay sensitive enough to capture important motion (0.3 for Human3.6M, 0.5 for MOT16 and WildTrack).
The resulting mask is then dilated by 7 pixels to also include smaller updates in the neighboring regions.
The effect of using large thresholds together with dilation on the input image compared to small thresholds without dilation is evaluated in Figure~\ref{fig:update_mask_mot16}.

\subsection{Human pose estimation}
For human pose estimation, we use two different CNN architectures: HRNet~\cite{Sun2019} and Pose-ResNet~\cite{Xiao2018}.
The networks are initialized with weights pre-trained on ImageNet \cite{Deng2010} and further trained on Human3.6M \cite{Ionescu2014} with an input resolution of $384$x$384$.
Human3.6M is designed as a benchmark and does not provide the ground truth poses for the test set publicly.
As our evaluation includes frame-by-frame analysis of accuracy, FLOPs and speedup, we use parts of the training set (\emph{Subject S11}) for testing, and exclude them during training. %, to be able to compare the predictions against the ground truth.
The test set contains 120 videos with an average length of 1927 frames and therefore serves as a reference for how much error accumulates over long time evaluations with \ours.

\subsection{Object detection}
For object detection, we use EfficientDet \cite{Tan2020}, based on the parameter- and FLOPs-efficient EfficientNet architecture \cite{Tan2019}.
The network is trained on two different datasets: Multiple Object Tracking 16 (MOT16) \cite{Dendorfer2021} and WildTrack \cite{Chavdarova2018}.
In both cases, we trained multiple configurations of EfficientDet on the video datasets with a 80/20 train/test split and initialized the network with weights pre-trained on the COCO dataset~\cite{Lin2014}.
WildTrack provides videos with a frame rate of 60 frames per second (FPS), but the provided ground truth annotations use a frame rate of only 2 FPS.
For speedup and accuracy evaluation, we feed the CNN with 60 FPS images to simulate real-time camera input, but report the accuracy only for every 30th frame.
Both datasets are recorded in a 16:9 aspect ratio.
Since EfficientDet is expecting an 1:1 input, we fill up the rest of the image with black pixels for accuracy evaluation.
However, in this case, 43\% of the image would never change and therefore lead to unfair advantage in performance comparisons.
We level the playing field by scaling up the image uniformly and performing center cropping for frame rate measurements.

\subsection{Hardware}
Performance evaluations were conducted on three devices with different power targets and from different hardware generations: 1) Jetson Nano: a low-end mobile development kit with a power target of 10W and 128 CUDA cores. 2) Dell XPS 9560: a notebook equipped with a Nvidia GTX 1050 with 640 CUDA cores. 3) Desktop PC: a high-end desktop PC equipped with a Nvidia RTX 3090 with 10496 CUDA cores.
% \begin{enumerate}
%     \item Jetson Nano: a low-end mobile development kit with a power target of 10W and 128 CUDA cores.
%     \item Dell XPS 9560: a notebook released in 2017 equipped with a Nvidia GTX 1050 with 640 CUDA cores.
%     \item Desktop PC: a high-end desktop PC equipped with a Nvidia RTX 3090 with 10496 CUDA cores.
% \end{enumerate}
The evaluations are performed with 32-bit floating point on the GTX 1050 and RTX 3090. 
On Jetson Nano, we use 16-bit floating point to reduce the memory overhead of weights and caches and to double the FLOPS throughput.

% \todo{Accuracy: PCK, (AP?)}

% \todo{Theoretical FLOPs: tile size = 1px}

% \todo{Actual FLOPs: real tile sizes + mostly dense tiles}

% \todo{Theoretical Speedup: FLOPs}

% \todo{Conv Speedup}

% \todo{Actual Speedup: with all overhead involved}

% \todo{Memory overhead}

% \todo{Memory bandwidth}
\section{Results}
\begin{figure}[pt]
\begin{subfigure}{.93\linewidth}
\begin{subfigure}{.25\linewidth}
  \centering
  \includegraphics[width=.98\linewidth]{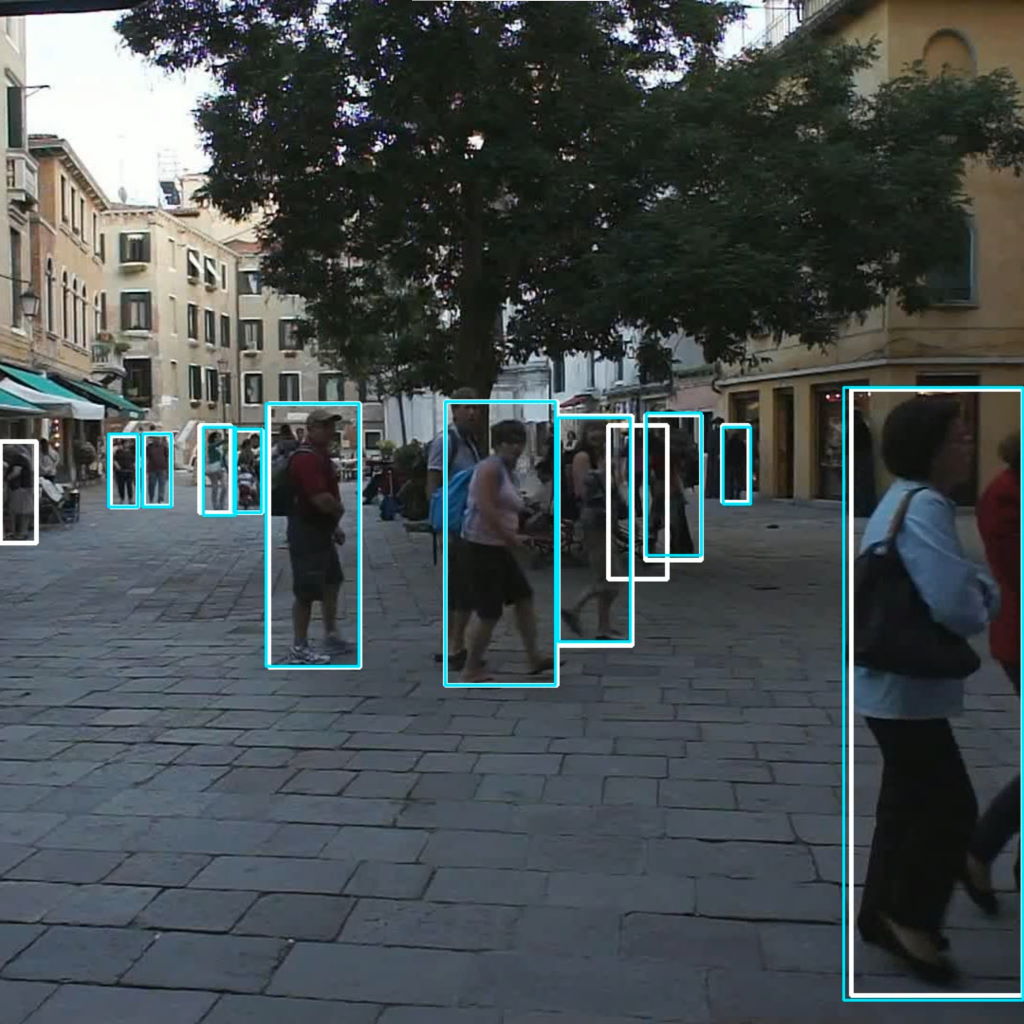}
\end{subfigure}%
\begin{subfigure}{.25\linewidth}
  \centering
  \includegraphics[width=.98\linewidth]{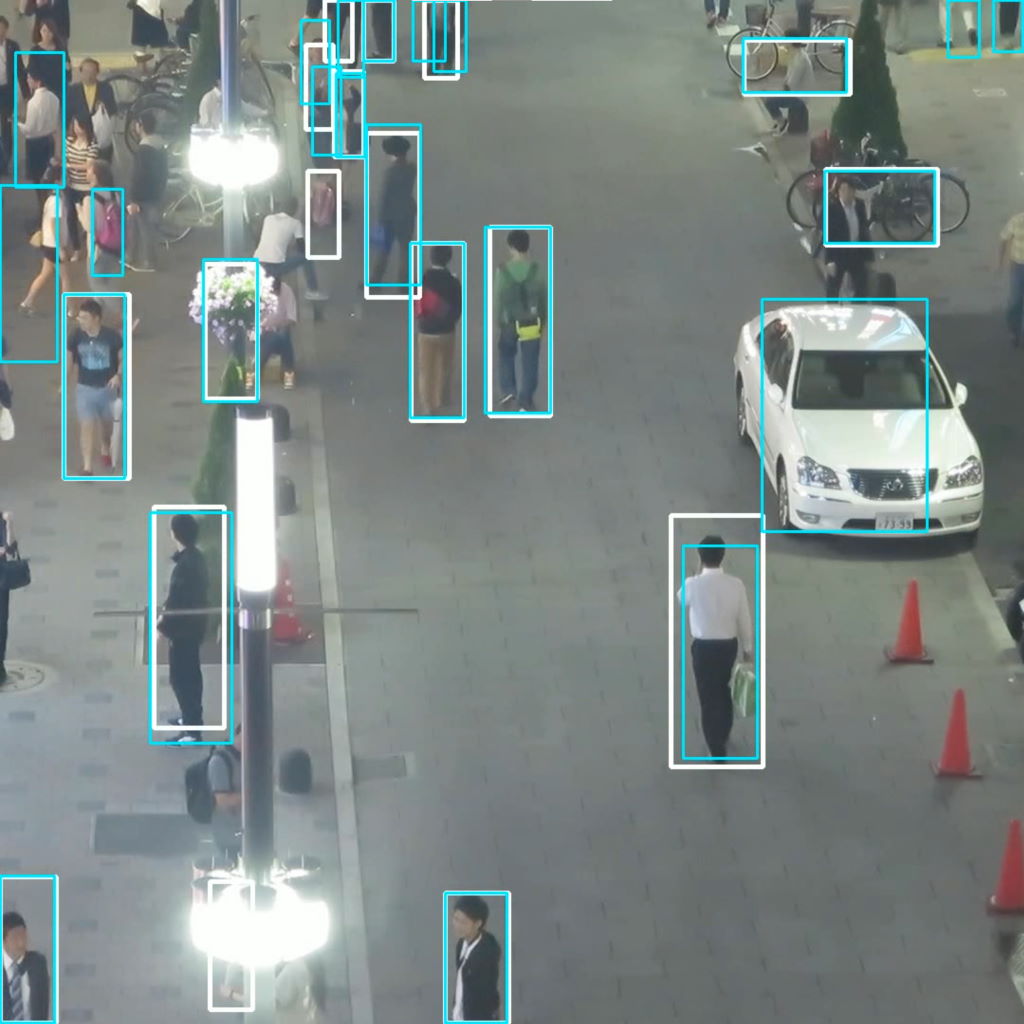}
\end{subfigure}%
\begin{subfigure}{0.25\linewidth}
  \centering
  \includegraphics[width=.98\linewidth]{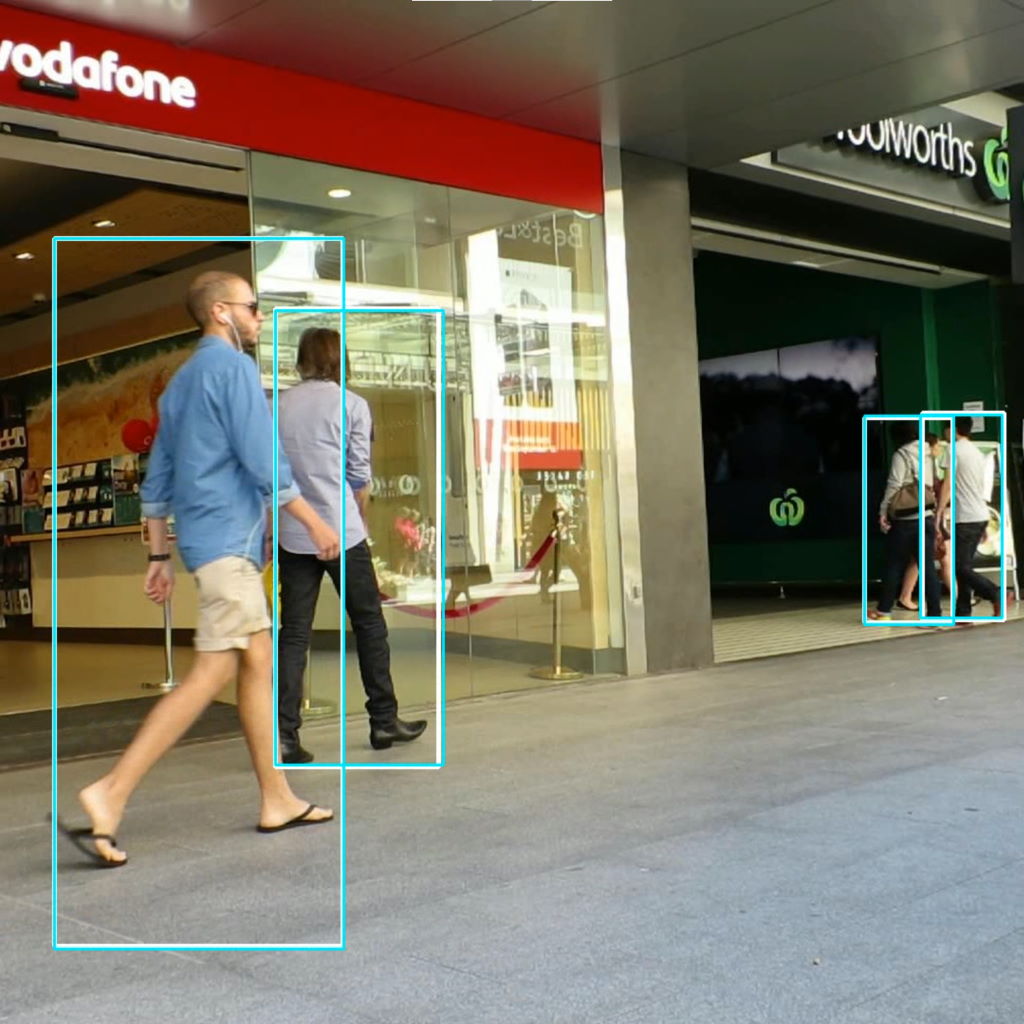}
\end{subfigure}%
\begin{subfigure}{0.25\linewidth}
  \centering
  \includegraphics[width=.98\linewidth]{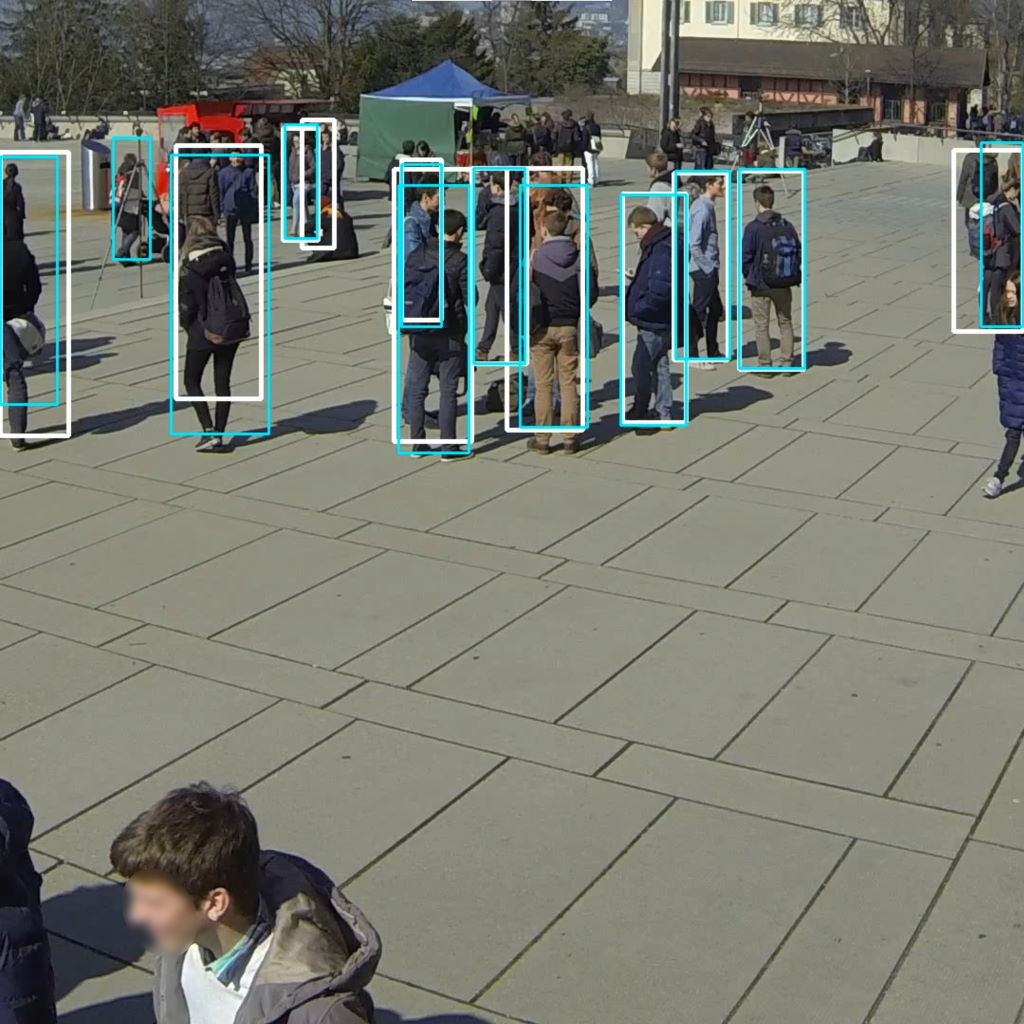}
\end{subfigure}%

\begin{subfigure}{.25\linewidth}
  \centering
  \includegraphics[width=.98\linewidth]{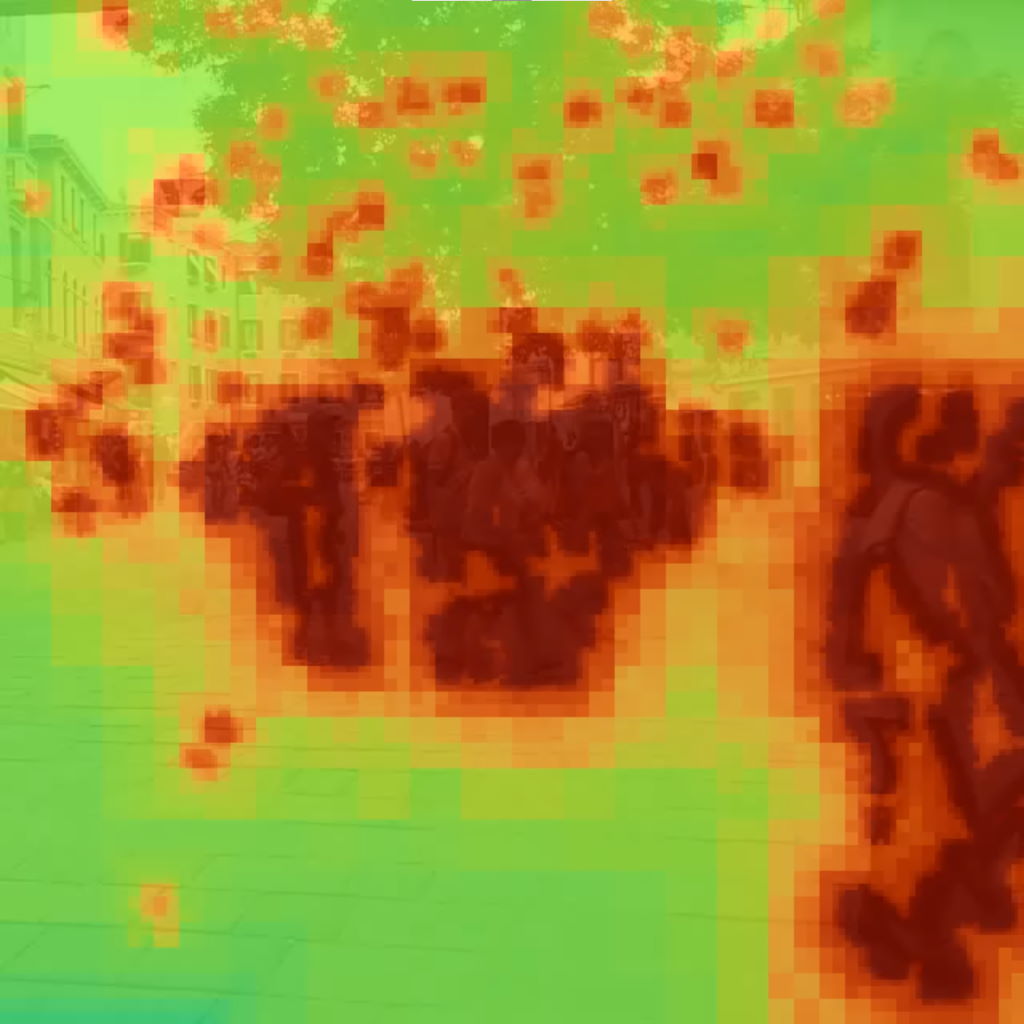}
\end{subfigure}%
\begin{subfigure}{.25\linewidth}
  \centering
  \includegraphics[width=.98\linewidth]{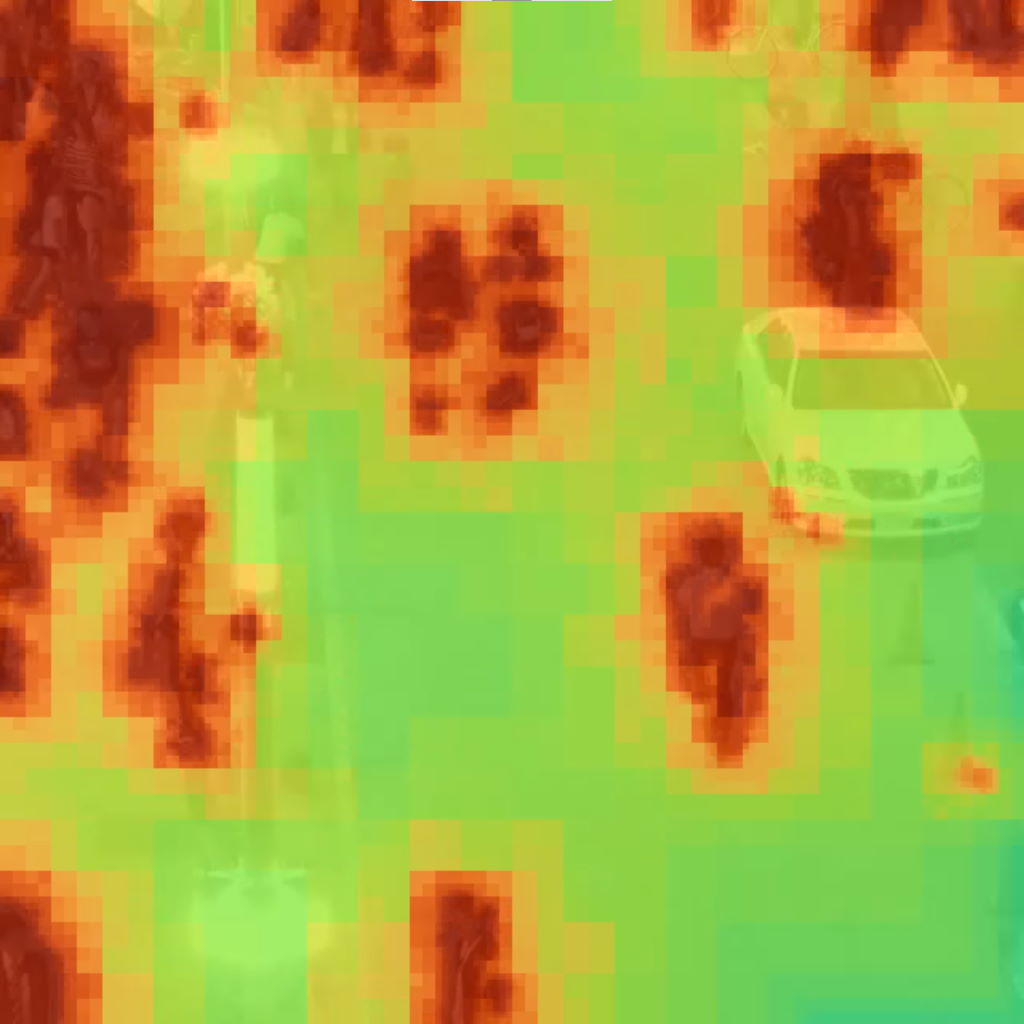}
\end{subfigure}%
\begin{subfigure}{0.25\linewidth}
  \centering
  \includegraphics[width=.98\linewidth]{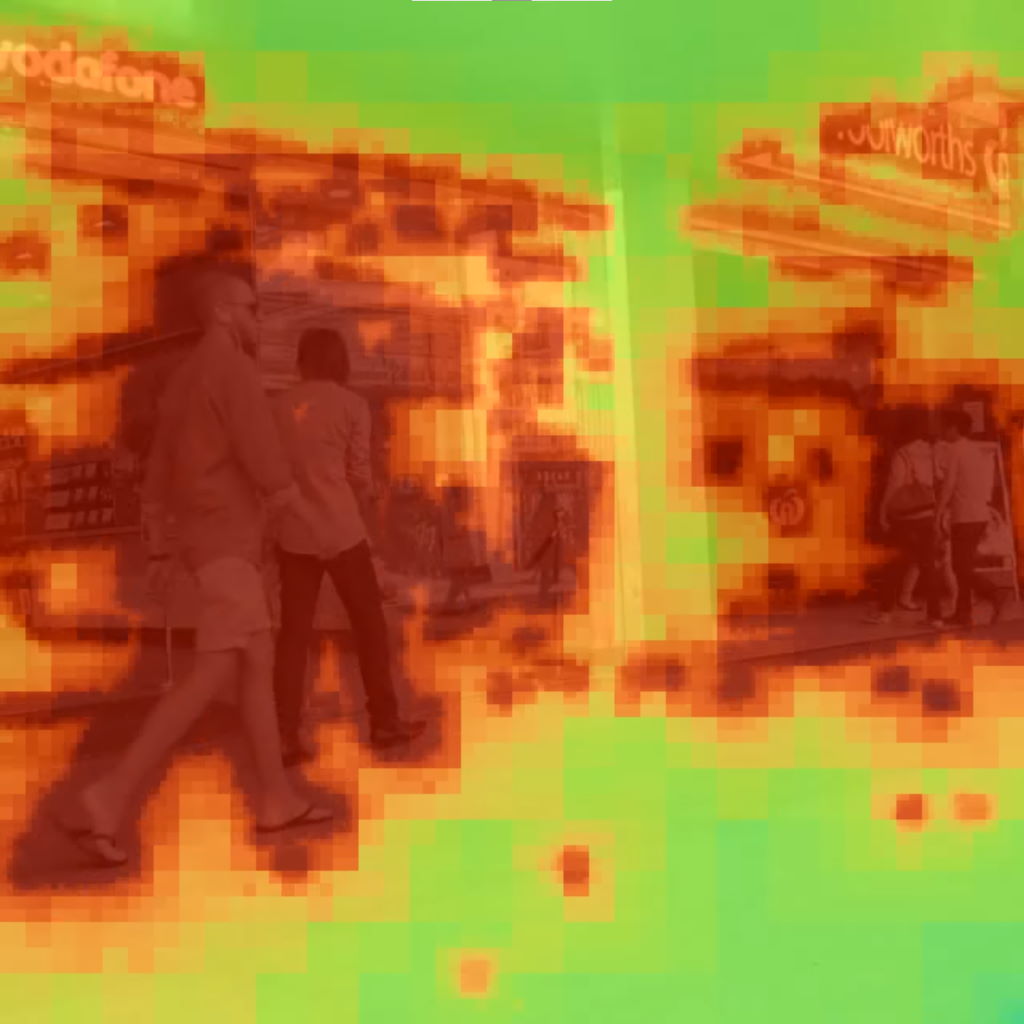}
\end{subfigure}%
\begin{subfigure}{0.25\linewidth}
  \centering
  \includegraphics[width=.98\linewidth]{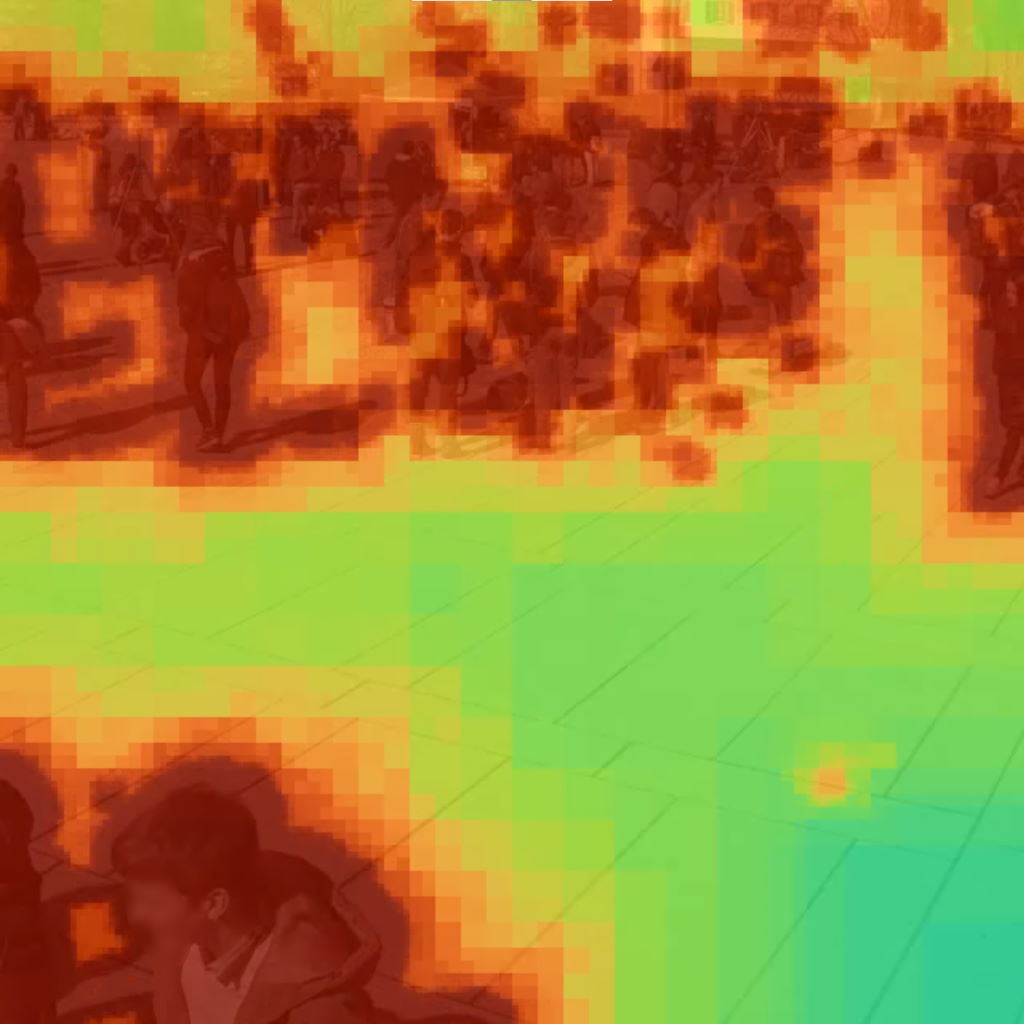}
\end{subfigure}%
\end{subfigure}%
% \hspace{0.01\linewidth}
\begin{subfigure}{.058\linewidth}
  \centering
  \includegraphics[width=\linewidth]{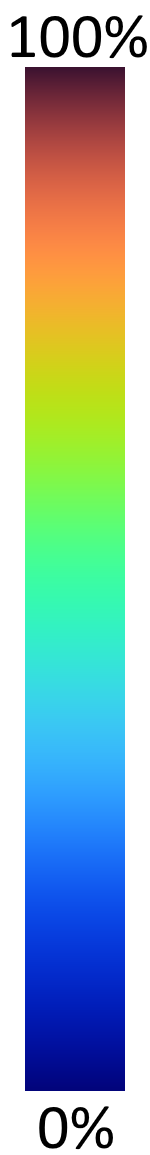}
\end{subfigure}

\caption{
Results using \ours to accelerate EfficientDet-d1 inference on MOT16 and WildTrack.
The first row shows predicted bounding boxes with \ours (blue) and with the dense reference network (white).
The second row shows how many of the convolutional layers updated each pixel. 
}
\label{fig:mot16_pred_update2}
\end{figure}
We evaluate the accuracy versus speedup trade-off for human pose estimation and object detection before analyzing \ours's characteristics and side effects of the sparse updates on temporal data. 

\subsection{Human pose estimation}

\label{sec:humanPoseEstimation}
For human pose estimation, we report accuracy as \emph{probability of correct keypoint normalized over head segment length (PCKh)} \cite{andriluka2014} and compare the throughput with different batch sizes.
Our results show that \ours~significantly speeds up inference compared to \emph{cuDNN} and previous work, with marginal loss of accuracy (Table \ref{tab:acc_speedup_pose_estimation}).
Tuning \emph{CBInfer}'s thresholds to exactly match the accuracy of our method is difficult and can only be approximated.
To emphasize \ours's speed advantage, we relaxed the accuracy requirement when tuning the thresholds for \emph{CBInfer} to allow it to leverage higher levels of sparsity.
Even enforcing a much higher accuracy, we still outperform \emph{CBInfer} speedwise by 3x.
It should be mentioned that due to \emph{CBInfer}'s lack of support for strided, dilated and depth-wise convolutions, 14\% of HRNet's and 7\% of Pose-Resnet's convolutions are processed with the dense \emph{cuDNN} backend.
% PoseResNet uses transposed convolutions for upscaling at the end of the network.
% Since we did not implement a sparse variant, we use the dense cuDNN implementation for the last 5 layers of the network and can therefore not achieve the same level of speedup.
% \todo{discuss PoseResNet and why it is more difficult to accelerate? I don't know yet why. Probably the thresholds are just too accurate still}
% \mathias{PoseResNet is now End-To-End Sparse. I am currently evaluating the improved speedup, but should be similar to before (around +15\%).}
% \todo{Add animation for update mask density \& output comparison}

We accelerate inference on all tiers of GPUs, with higher speedup on slower devices.
High-end GPUs like the RTX 3090 require larger batch sizes to benefit from skipping single tiles.
This is because the RTX 3090 has over 10000 compute cores compared to 128 on the Jetson Nano.
A batch size of one is typically not large enough to fully utilize the GPU when processing a single low-resolution input image.
% In this case, skipping some tiles does not have an impact on performance, because the freed up resources cannot be used elsewhere, as they could with increased workload using larger batch sizes.
The surplus of available compute resources for these GPUs means freeing them up by skipping tiles makes little difference.

Detailed evaluations over all frames and kernels for HRNet show that only 6\% of input pixels in convolutional layers are updated on average in our approach.
16\% of the tiles were processed, resulting in a 84\% reduction in FLOPs, and the overall memory transfers are reduced to 21\% despite the overhead of reading and updating the additional buffers.

\subsection{Object detection}
For object detection, we use the Average Precision (AP) metric to measure accuracy.
Compared to HRNet and Human3.6M, the EfficientDet model is computationally lightweight and the two datasets contain much denser motion (see Figure \ref{fig:mot16_pred_update2}).
Combined, this makes it more difficult to speed up compared to a heavy network and small and centered frame updates.
Still, we are able to accelerate inference by a factor of 2.5x to 6.7x over the \emph{cuDNN} baseline on both datasets and all devices with nearly identical accuracy.

Because \emph{CBInfer} lacks support for strided and depthwise convolutions (Section~\ref{sec:humanPoseEstimation}), a third of the layers are processed with the \emph{cuDNN} backend.
\emph{CBInfer} struggles to accelerate the lightweight network compared to the dense version and is in some cases over 2x slower than \emph{cuDNN}, even with only 2/3rds of the layers replaced and a 65\% reduction in FLOPs.

EfficientDet and EfficientNet provide multiple configurations of the networks (d0-d7) that use different resolutions, layer counts and numbers of channels.
This allows us to show that \ours does not always require trading accuracy for speedup.
Instead of using the \emph{cuDNN} backend for the \emph{d0} configuration, \ours allows for much more accurate predictions (AP@0.5 of 64.0\% vs 55.6\%  on the MOT16 dataset) with the \emph{d1} configuration with slightly higher frame rates (+5\% on RTX 3090 up to +279\% on Jetson Nano).

On the MOT16 dataset, we process nearly 60\% of FLOPs of the dense baseline.
However, on average, we are able to skip over 60\% of the tiles, reducing memory bandwidth by 60\%, and thereby achieve over 2x speedup over the dense baseline. 
% \todo{interesting comparison against wildtrack: both have nearly identical ratio of active tiles, but wildtrack requires only 49\% of dense FLOPs and is also only slightly faster}
Detailed evaluations of accuracy and frame rates for both datasets and both network configurations are reported in the supplementary material.

\begin{table*}[ptb]
\centering
\setlength{\tabcolsep}{4pt}
\renewcommand\arraystretch{1.2}
\resizebox{\linewidth}{!}{
\begin{tabular}{cc||c|c|c|cc|cc|cc|cc|cc}
\hline\thickhline
\multirow{2}{*}{CNN} & \multirow{2}{*}{Backend} & \multirow{2}{*}{PCKh@0.5} & \multirow{2}{*}{PCKh@0.2} & \multirow{2}{*}{GFLOPs} & \multicolumn{2}{c|}{Jetson Nano} & \multicolumn{2}{c|}{GTX 1050 b=1} & \multicolumn{2}{c|}{GTX 1050 b=4} & \multicolumn{2}{c|}{RTX 3090 b=1} & \multicolumn{2}{c}{RTX 3090 b=32} \\
\cline{6-15}
& & & & & FPS & speedup & FPS & speedup & FPS & speedup & FPS & speedup & FPS & speedup \\
\hline
\multirow{3}{*}{HRNet} 
& cuDNN         & \multirow{2}{*}{97.29\%}   & \multirow{2}{*}{87.25\%} & \multirow{2}{*}{47.1}
                                                                & 0.7   & 1.0   & 4.7   & 1.0   & 5.2   & 1.0   & 10.1  & 1.0   & 105   & 1.0 \\
& ours dense                &           &               &       & 1.1   & 1.5   & 6.8   & 1.4   & 7.1   & 1.4   & 26.9  & 2.7   & 97.6  & 0.9 \\
& ours $\epsilon=\infty$    & 28.07\%   & 13.88\%       & -     & 6.7   & 9.6   & 30.6  & 6.5   & 93.7  & 18.0  & 31.8  & 3.1   & 949   & 9.0   \\
& CBInfer                   & 96.94\%   & 85.00\%       & 13.9     & 1.5   & 2.1   & 4.9   & 1.0   & 10.7  & 2.1   & 6.7   & 0.7   & 125   & 1.2 \\
% & ours sparse 1             & 97.20\%   & 85.89\%       & 5.1   & 7.0   & 21.5  & 4.6   & 31.6  & 6.1   & 33.1  & 3.3   & 439   & 4.2 \\
% & ours sparse 2             & 97.27\%   & 86.33\%       & -     & -     & 20.1  & 4.3   & 26.5  & 5.1   & -     & -     & 433   & 4.1 \\
& ours sparse               & 97.27\%   & 86.33\%       & 7.7   & 4.7   & 6.7   & 20.1  & 4.3   & 26.5  & 5.1   & 30.5  & 3.0   & 433   & 4.1 \\
% & ours sparse trained 1     & -         & 87.18\%       & -     & -     & -     & -     & -     & -     & -     & -     & \todo{ca. 305}  & 2.9 \\
% & ours sparse trained 2     & 97.35         & 86.79\%       & -     & -     & -     & -     & -     & -     & -     & -     & 346  & 3.3 \\
\hline
\multirow{3}{*}{ResNet} 
& cuDNN         & \multirow{2}{*}{95.78\%}   & \multirow{2}{*}{82.79\%} & \multirow{2}{*}{27.2}
                                                        &  1.5  & 1.0   & 7.7   & 1.0   & 10.4  & 1.0   & 30.4  & 1.0   & 215   & 1.0 \\
& ours dense                &           &               && 1.7  & 1.1   & 9.2   & 1.2   & 9.8   & 0.9   & 63.3  & 2.1   & 187   & 0.9 \\
& ours $\epsilon=\infty$    & 27.97\%   & 13.77\%       & -     & 13.4 & 8.9   & 30.8  & 4.0   & 42.5  & 4.1   & 67.6  & 2.2   & 1838  & 8.5 \\
% & ours sparse 1 & 95.49\%   & 81.27\%       & -     & -     & 20.5  & 2.6   & 22.5  & 2.2   & 63.3  & 2.1   & 497   & 2.3 \\
% & ours sparse 2 & -         & 82.68\%       & -     & -     & -     & -     & -     & -     & -     & -     & \todo{ca. 540}& 2.5 \\
% & CBInfer                   & -         & 82.73\%       & -     & -     & 6.6   & -     & 15.9  & -     & -     & -     & -     & - \\
& CBInfer                   & 95.82\%   & 82.68\%       & 17.6  & 2.6     & 1.7     & 5.6   & 0.7   & 16.6  & 1.6   & 16.6  & 0.5   & 236 & 1.1 \\
& ours sparse               & 95.82\%   & 82.68\%       & 11.6  & 5.7   & 3.8   & 20.5  & 2.7   & 27.5  & 2.6   & 67.4  & 2.2   & 577 & 2.7 \\
\hline\thickhline
\end{tabular}
}
\caption{Speed and accuracy comparisons of different CNN backends used for pose estimation on the Human3.6M dataset. The same set of auto-tuned thresholds for update truncation is used for all devices and batch sizes $b$.
}
\label{tab:acc_speedup_pose_estimation}
\end{table*}

\subsection{Additional evaluations}
Besides differences in speed and accuracy, temporal reuse and update truncation cause side effects.

\begin{figure}[pt]
% dense
% \begin{subfigure}{0.25\linewidth}
% Dense
% \end{subfigure}%
% \begin{subfigure}{.25\linewidth}
%   \centering
%   \includegraphics[width=.99\linewidth]{figures/pose_stability/dense3.png}
% \end{subfigure}%
\begin{subfigure}{.25\linewidth}
  \centering
  \includegraphics[width=.99\linewidth]{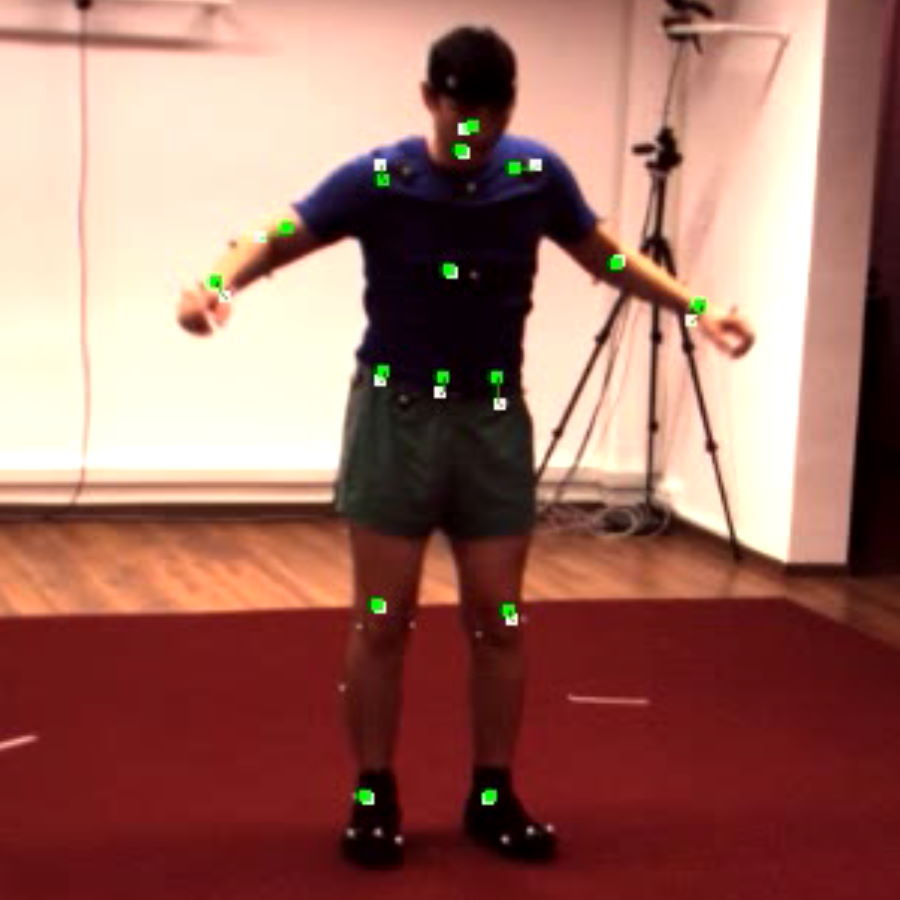}
  \caption{$I^{1}$ dense / ours}
\end{subfigure}%
\begin{subfigure}{.25\linewidth}
  \centering
  \includegraphics[width=.99\linewidth]{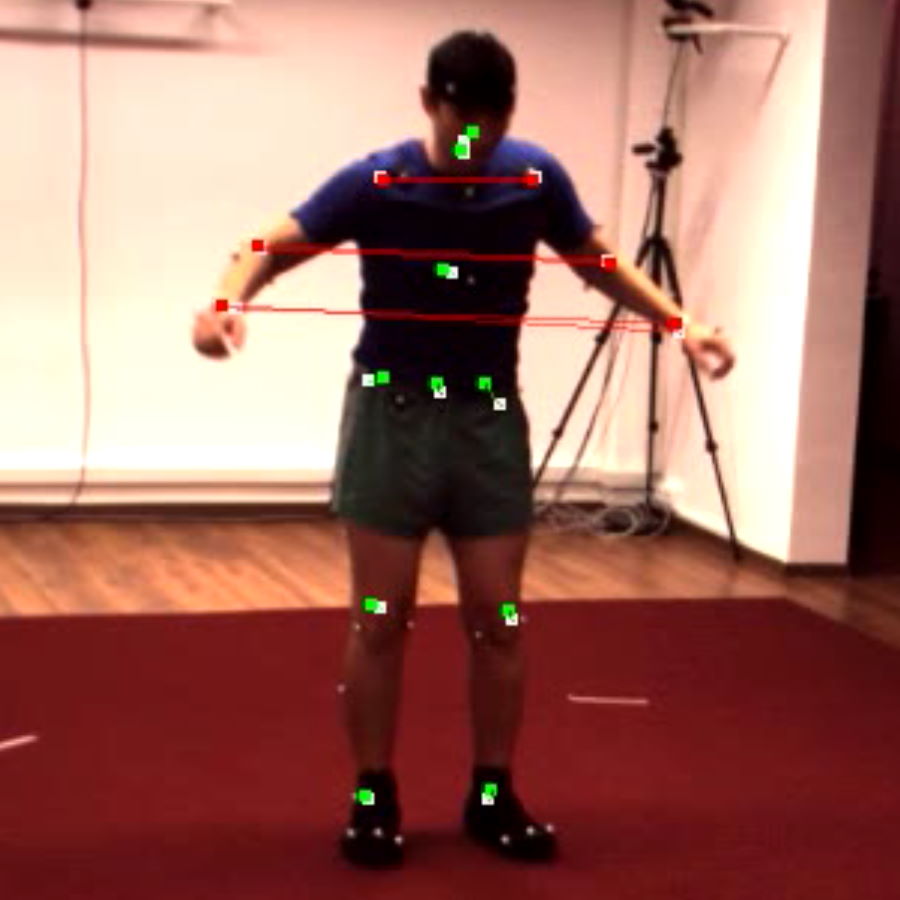}
  \caption{$I^{2}$ dense}
\end{subfigure}%
% \begin{subfigure}{0.33\linewidth}
%   \centering
%   \includegraphics[width=.99\linewidth]{figures/pose_stability/dense1.png}
% \end{subfigure}%
% sparse
% \begin{subfigure}{0.25\linewidth}
% \ours
% \end{subfigure}%
% \begin{subfigure}{.25\linewidth}
%   \centering
%   \includegraphics[width=.99\linewidth]{figures/pose_stability/sparse3.png}
% \end{subfigure}%
% \begin{subfigure}{.25\linewidth}
%   \centering
%   \includegraphics[width=.99\linewidth]{figures/pose_stability/sparse2.png}
% \end{subfigure}%
\begin{subfigure}{0.25\linewidth}
  \centering
  \includegraphics[width=.99\linewidth]{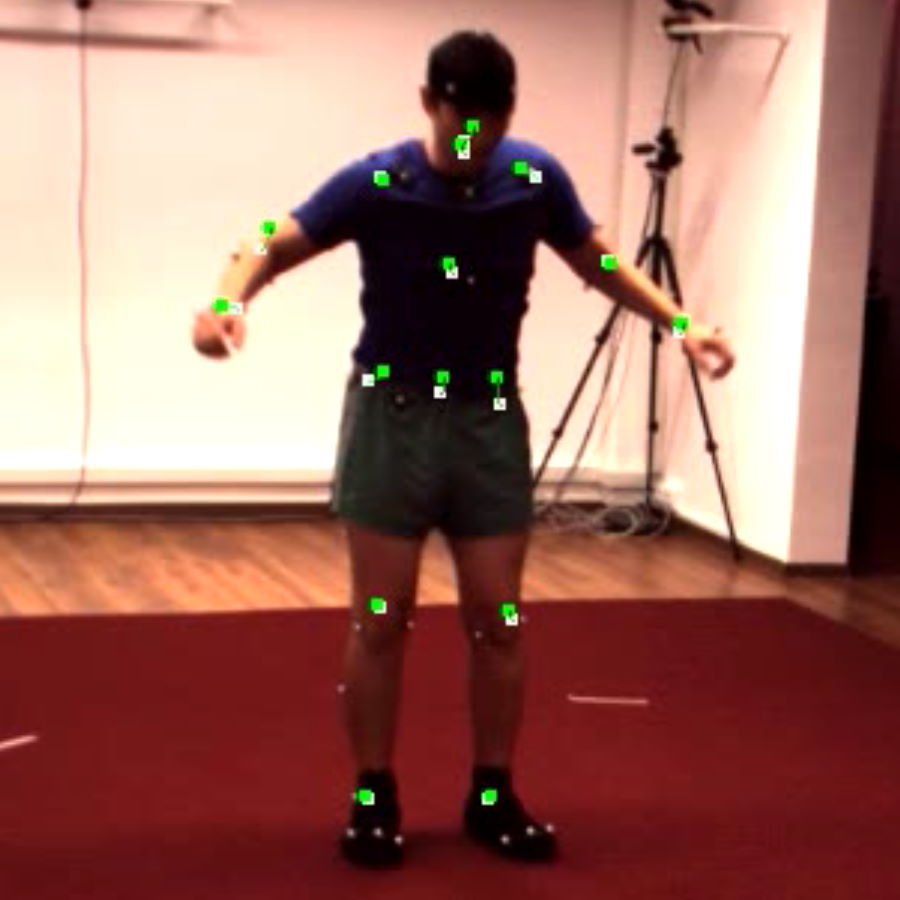}
  \caption{$I^{2}$ ours}
\end{subfigure}%
\begin{subfigure}{0.25\linewidth}
  \centering
  \includegraphics[width=.99\linewidth]{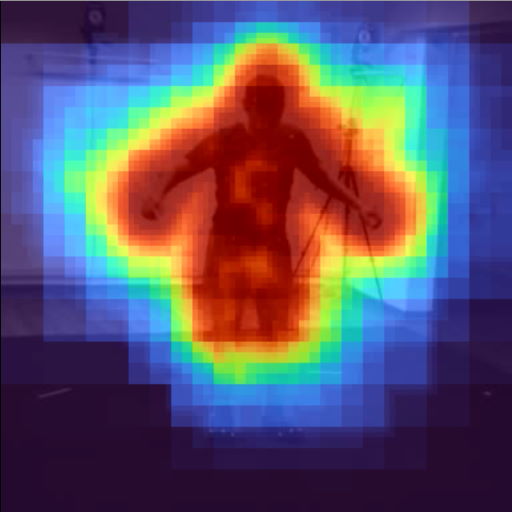}
  \caption{$I^{2}$ updates}
  \label{img:pose_stability:updates}
\end{subfigure}%

\caption{\ours~can improve tracking accuracy in some cases due to temporal stability. 
Here, the dense reference confused left and right arms in frame $I^{2}$, whereas \ours~predicted all joint positions correctly by reusing cached information from $I^{1}$. 
White, green and red points indicate ground truth, correct and incorrect predictions respectively.
Image (d) shows how often a region was updated in $I^{2}$ on average over all HRNet layers using \ours.
}
\label{fig:pose_stability}
\end{figure}
\textbf{Improved stability}
When we set the goal of minimizing the number of updates, predictions become more temporally stable and react less to noise in the camera image or by overly sensitive networks.
For example, for pose estimation with HRNet, the dense baseline produces 43\% more joint position updates compared to \ours.
% dense jitter: 0.2265, our jitter: 0.1578. Both after 44k frames out of 230k, I think that should be enough
This effect is especially visible when joints are static, but the dense prediction constantly jitters positions by a few pixels whereas \ours~stays constant (see feet in Figure \ref{fig:pose_stability}).
The temporal stability can even help resolving difficult-to-track poses.
Tiny frame-to-frame differences can lead to very different results with dense inference, whereas \ours mostly relies on information from previous frames and only applies sparse updates (Figure \ref{fig:pose_stability}).

\textbf{Threshold analysis}
Analyzing the tuned thresholds can reveal interesting insights about the contribution of each layer, or about how sensitive layers react to updates on the input.
In case of EfficientDet, some branches of the EfficientNet backbone are turned off completely after the first frame, because the frame-by-frame changes do not impact the result.
EfficientNet features \emph{squeeze and excitation} layers, which use average pooling to scale the feature map down to a single pixel.
Two convolutions and two activation functions are applied to this pixel before it is used to scale the features of the original feature map.
Since average pooling over a sequence of a few seconds returns nearly identical results for all frames, all updates are truncated in the first activation/truncation layer in the squeeze and excitation branch.

\textbf{Overhead}
Table \ref{tab:acc_speedup_pose_estimation} shows that \ours comes with very low overhead.
In dense mode, we use negative thresholds to guarantee that all pixels will be reprocessed every time, allowing to compare against \emph{cuDNN}.
Like in sparse mode, we use \emph{delta} values, accumulated and truncated values buffers and perform all steps including truncation to include all overhead. 
In most cases, especially for smaller batch sizes, we still outperform \emph{cuDNN} even when taking the extra steps.
This may in part be due to static overheads and different parallelization strategies, leading to \ours providing more utilization on large GPUs. 
%This is in parts caused by the internal overhead of PyTorch which becomes more noticeable with smaller batch sizes.
Yet, these evaluations show that \ours does not require a minimum level of sparsity to reach the break-even point between overhead and gains.

As \ours stores accumulated and truncated values, memory overhead scales linearly with batch size and is larger than the memory needed for evaluating the dense version.
Recall that memory \emph{bandwidth} is still reduced by \ours compared to dense evaluation.
Depending on the network architecture, storage overhead can reduce gains when the largest possible batch size is too small to fully utilize the GPU.
At the same time, \ours reduces cache size compared to \emph{CBInfer} by 28\% and 13\% with HRNet and EfficientDet.
% \todo{talk about memory: we buffer 115.8M x batch size values in HRNet, CBInfer buffers 160.7M values. EfficientDet-d0 71.3M vs 81.3M, EfficientDet-d1 153.4M vs 179.0M}

% \todo{mention: in dense mode, we read 15.7\% more values and write 27.6\% more values due to additional buffers in EfficientDet-d1. }

% \todo{add table with accuracy numbers for different configurations}
% \todo{add table with FLOPs \& FPS for different configurations}
% \todo{evaluate memory overhead \& max batch sizes}
% \todo {evaluate memory bandwidth where possible}

% \include{figures/update_masks/update_masks_walk_figure}

\section{Discussion}
Our evaluations show that \ours can speedup video inference with marginal loss of accuracy.
Since \ours comes with significantly less overhead than previous work, our implementation can accelerate expensive CNNs just like FLOPs-efficient CNNs, datasets with sparse or dense updates and low-end as well as high-end GPUs alike.
Thresholds can be tuned to gain identical accuracy with small speedup, or small decreases of accuracy with large speedup.
\ours can even increase the accuracy at the same frame rate by allowing the use of larger networks.

\textbf{Tiled convolution}
Our convolutions are processed in tiles, with a single pixel update on the input causing an entire tile to be processed.
Compared to \emph{CBInfer}, this reduces the computational savings for unstructured sparsity, as \emph{CBInfer} can control sparsity per output pixel.
Yet, as the authors of \emph{CBInfer} stated, frame-to-frame updates are typically structured and per pixel sparsity mainly helps to accelerate the halo of updated regions \cite{Cavigelli2020}, which contribute only a small portion of the updated pixels.
At the same time, our approach achieves state-of-the-art performance for dense inference, allowing us to accelerate even very dense scenes from the MOT16 and WildTrack datasets.

\textbf{Limitations}
One major limitation of temporally sparse CNNs, and therefore also of \ours, is that they only work well on fixed camera input.
Even small camera motion can lead to nearly dense updates, at least for the first few layers.
Later layers with lower resolution input are often able to truncate parts of the updates, but overall speedup still deteriorates.

Another disadvantage of temporally sparse CNNs is the memory overhead which increases linearly with the batch size.
Compared to \emph{RRM}, \emph{Skip-Convolutions} and \emph{CBInfer}, \ours requires roughly 20\% less cache memory. 
Still, memory can be a limiting factor, especially on low-end devices.
If device memory is insufficient, memory overhead could be lowered by omitting \emph{delta} truncation on some of the activation layers.
In this case, the truncated values buffer $x^T$ can be avoided at the cost of greater update density.

% \todo{talk about threshold training. I don't want to optimize all thresholds again, that takes too long. But I do have some results showing that the accuracy/speedup trade off that we achieved is very similar to tuned thresholds, but it takes much longer and it is way more difficult to get the parameters right. So we recommend simple tuning instead. Also, max-truncation is faster than norm-truncation, because we don't need to load all values twice everytime (when one of the first 32 values is > threshold, we don't need to check other values)}

\section{Conclusion}
\label{sec:conclusion}
In this paper, we describe \ours, a method and the corresponding implementation for accelerating CNN inference on video input.
\ours is, to the best of our knowledge, the first solution to offer an end-to-end propagation of sparse frame updates and the first approach of its kind to achieve speedups in practice.
With a design optimized for GPUs, we are able to outperform the state-of-the-art framework for dense inference and existing sparse implementations alike.
Our approach can be ported to other platforms and processors optimized for CNN acceleration, as the core of the convolution is unaware of sparsity, and speedup is gained by skipping tiles entirely without introducing large processing overhead.

% \todo{discuss how to adapt this approach for moving cameras, as mentioned in 2.1}

% this references style is copied over from the official template
{\small
\bibliographystyle{ieee_fullname}
\bibliography{references}

\begin{thebibliography}{10}\itemsep=-1pt

\bibitem{Alwis2021}
Udari~De Alwis and Massimo Alioto.
\newblock {TempDiff: Temporal Difference-Based Feature Map-Level Sparsity
  Induction in CNNs with \textless 4\% Memory Overhead}.
\newblock {\em 2021 IEEE 3rd International Conference on Artificial
  Intelligence Circuits and Systems, AICAS 2021}, pages 1--4, Jun 2021.

\bibitem{andriluka2014}
Mykhaylo Andriluka, Leonid Pishchulin, Peter Gehler, and Bernt Schiele.
\newblock {2D human pose estimation: New benchmark and state of the art
  analysis}.
\newblock {\em Proceedings of the IEEE Computer Society Conference on Computer
  Vision and Pattern Recognition}, pages 3686--3693, 2014.

\bibitem{Cavigelli2020}
Lukas Cavigelli and Luca Benini.
\newblock {CBinfer: Exploiting Frame-to-Frame Locality for Faster Convolutional
  Network Inference on Video Streams}.
\newblock {\em IEEE Transactions on Circuits and Systems for Video Technology},
  30(5):1451--1465, May 2020.

\bibitem{Chavdarova2018}
Tatjana Chavdarova, Pierre Baque, Stephane Bouquet, Andrii Maksai, Cijo Jose,
  Timur Bagautdinov, Louis Lettry, Pascal Fua, Luc {Van Gool}, and Francois
  Fleuret.
\newblock {WILDTRACK: A Multi-camera HD Dataset for Dense Unscripted Pedestrian
  Detection}.
\newblock {\em Proceedings of the IEEE Computer Society Conference on Computer
  Vision and Pattern Recognition}, pages 5030--5039, 2018.

\bibitem{Chen2016}
Yunji Chen, Tianshi Chen, Zhiwei Xu, Ninghui Sun, and Olivier Temam.
\newblock {DianNao family: Energy-efficient hardware accelerators for machine
  learning}.
\newblock {\em Communications of the ACM}, 59(11):105--112, 2016.

\bibitem{Chen2015}
Yunji Chen, Tao Luo, Shaoli Liu, Shijin Zhang, Liqiang He, Jia Wang, Ling Li,
  Tianshi Chen, Zhiwei Xu, Ninghui Sun, and Olivier Temam.
\newblock {DaDianNao: A Machine-Learning Supercomputer}.
\newblock {\em Proceedings of the Annual International Symposium on
  Microarchitecture, MICRO}, 2015-January:609--622, Jan 2015.

\bibitem{Dendorfer2021}
Patrick Dendorfer, Aljosa Osep, Anton Milan, Konrad Schindler, Daniel Cremers,
  Ian Reid, Stefan Roth, and Laura Leal-Taix{\'{e}}.
\newblock {MOTChallenge: A Benchmark for Single-Camera Multiple Target
  Tracking}.
\newblock {\em International Journal of Computer Vision}, 129(4):845--881,
  2021.

\bibitem{Deng2010}
Jia Deng, Wei Dong, Richard Socher, Li-Jia Li, {Kai Li}, and {Li Fei-Fei}.
\newblock {ImageNet: A large-scale hierarchical image database}.
\newblock pages 248--255, Mar 2010.

\bibitem{Fan2020}
Zhipeng Fan, Jun Liu, and Yao Wang.
\newblock {Adaptive Computationally Efficient Network for Monocular 3D Hand
  Pose Estimation}.
\newblock Technical report, 2020.

\bibitem{Graham2018}
Benjamin Graham, Martin Engelcke, and Laurens {Van Der Maaten}.
\newblock {3D Semantic Segmentation with Submanifold Sparse Convolutional
  Networks}.
\newblock {\em Proceedings of the IEEE Computer Society Conference on Computer
  Vision and Pattern Recognition}, pages 9224--9232, Nov 2018.

\bibitem{Habibian_2021_CVPR}
Amirhossein Habibian, Davide Abati, Taco~S. Cohen, and Babak~Ehteshami
  Bejnordi.
\newblock {Skip-Convolutions for Efficient Video Processing}.
\newblock In {\em Proceedings of the IEEE/CVF Conference on Computer Vision and
  Pattern Recognition (CVPR)}, pages 2695--2704, Jun 2021.

\bibitem{Han2016}
Song Han, Xingyu Liu, Huizi Mao, Jing Pu, Ardavan Pedram, Mark~A. Horowitz, and
  William~J. Dally.
\newblock {EIE: Efficient Inference Engine on Compressed Deep Neural Network}.
\newblock {\em Proceedings - 2016 43rd International Symposium on Computer
  Architecture, ISCA 2016}, pages 243--254, 2016.

\bibitem{Han2015}
Song Han, Jeff Pool, John Tran, and William~J. Dally.
\newblock {Learning both weights and connections for efficient neural
  networks}.
\newblock {\em Advances in Neural Information Processing Systems},
  2015-January:1135--1143, 2015.

\bibitem{Hubara2018}
Itay Hubara, Matthieu Courbariaux, Daniel Soudry, Ran El-Yaniv, and Yoshua
  Bengio.
\newblock {Quantized neural networks: Training neural networks with low
  precision weights and activations}.
\newblock {\em Journal of Machine Learning Research}, 18:1--30, 2018.

\bibitem{Ionescu2014}
Catalin Ionescu, Dragos Papava, Vlad Olaru, and Cristian Sminchisescu.
\newblock {Human3.6M: Large scale datasets and predictive methods for 3D human
  sensing in natural environments}.
\newblock Technical Report~7, 2014.

\bibitem{Jain2019}
Samvit Jain, Xin Wang, and Joseph~E. Gonzalez.
\newblock {Accel: A corrective fusion network for efficient semantic
  segmentation on video}.
\newblock {\em Proceedings of the IEEE Computer Society Conference on Computer
  Vision and Pattern Recognition}, 2019-June:8858--8867, Jun 2019.

\bibitem{Kumar2019}
Athindran~Ramesh Kumar, Balaraman Ravindran, and Anand Raghunathan.
\newblock {Pack and detect: Fast object detection in videos using
  region-of-interest packing}.
\newblock {\em ACM International Conference Proceeding Series}, pages 150--156,
  2019.

\bibitem{Le1990}
CUN Le.
\newblock {Optimal brain damage}.
\newblock {\em Advances in Neural Information Processing Systems}, 2:598--605,
  1990.

\bibitem{Li2018}
Yule Li, Jianping Shi, and Dahua Lin.
\newblock {Low-Latency Video Semantic Segmentation}.
\newblock {\em Proceedings of the IEEE Computer Society Conference on Computer
  Vision and Pattern Recognition}, pages 5997--6005, 2018.

\bibitem{Lin2016}
Darryl~D. Lin, Sachin~S. Talathi, and V.~Sreekanth Annapureddy.
\newblock {Fixed point quantization of deep convolutional networks}.
\newblock {\em 33rd International Conference on Machine Learning, ICML 2016},
  6:4166--4175, 2016.

\bibitem{Lin2014}
Tsung~Yi Lin, Michael Maire, Serge Belongie, James Hays, Pietro Perona, Deva
  Ramanan, Piotr Doll{\'{a}}r, and C.~Lawrence Zitnick.
\newblock {Microsoft COCO: Common objects in context}.
\newblock {\em Lecture Notes in Computer Science}, 8693 LNCS(PART 5):740--755,
  2014.

\bibitem{Moons2018}
Bert Moons, Koen Goetschalckx, Nick {Van Berckelaer}, and Marian Verhelst.
\newblock {Minimum energy quantized neural networks}.
\newblock {\em Conference Record of 51st Asilomar Conference on Signals,
  Systems and Computers, ACSSC 2017}, 2017-October:1921--1925, Apr 2018.

\bibitem{Nie2019}
Xuecheng Nie, Yuncheng Li, Linjie Luo, Ning Zhang, and Jiashi Feng.
\newblock {Dynamic kernel distillation for efficient pose estimation in
  videos}.
\newblock {\em Proceedings of the IEEE International Conference on Computer
  Vision}, 2019-October:6941--6949, Oct 2019.

\bibitem{Pan2018}
Bowen Pan, Wuwei Lin, Xiaolin Fang, Chaoqin Huang, Bolei Zhou, and Cewu Lu.
\newblock {Recurrent Residual Module for Fast Inference in Videos}.
\newblock Technical report, 2018.

\bibitem{Peng_2021_WACV}
Gao Peng, Bo Pang, and Cewu Lu.
\newblock {Efficient 3D Video Engine Using Frame Redundancy}.
\newblock In {\em Proceedings of the IEEE/CVF Winter Conference on Applications
  of Computer Vision (WACV)}, pages 3792--3802, Jan 2021.

\bibitem{Ren2018}
Mengye Ren, Andrei Pokrovsky, Bin Yang, and Raquel Urtasun.
\newblock {SBNet: Sparse Blocks Network for Fast Inference}.
\newblock Technical report, 2018.

\bibitem{Shelhamer2016}
Evan Shelhamer, Kate Rakelly, Judy Hoffman, and Trevor Darrell.
\newblock {Clockwork convnets for video semantic segmentation}.
\newblock {\em Lecture Notes in Computer Science (including subseries Lecture
  Notes in Artificial Intelligence and Lecture Notes in Bioinformatics)}, 9915
  LNCS:852--868, 2016.

\bibitem{Sifre2014}
Laurent Sifre and St{\'{e}}phane Mallat.
\newblock {PhD Thesis, Ecole Polytechnique, CMAP Rigid-Motion Scattering For
  Image Classification}.
\newblock 2014.

\bibitem{Sun2019}
Ke Sun, Bin Xiao, Dong Liu, and Jingdong Wang.
\newblock {Deep high-resolution representation learning for human pose
  estimation}.
\newblock {\em Proceedings of the IEEE Computer Society Conference on Computer
  Vision and Pattern Recognition}, 2019-June:5686--5696, 2019.

\bibitem{Tan2019}
Mingxing Tan and Quoc~V. Le.
\newblock {EfficientNet: Rethinking model scaling for convolutional neural
  networks}.
\newblock Technical report, 2019.

\bibitem{Tan2020}
Mingxing Tan, Ruoming Pang, and Quoc~V. Le.
\newblock {EfficientDet: Scalable and efficient object detection}.
\newblock {\em Proceedings of the IEEE Computer Society Conference on Computer
  Vision and Pattern Recognition}, pages 10778--10787, 2020.

\bibitem{Xiao2018}
Bin Xiao, Haiping Wu, and Yichen Wei.
\newblock {Simple baselines for human pose estimation and tracking}, 2018.

\bibitem{Zhu2018}
Xizhou Zhu, Jifeng Dai, Lu Yuan, and Yichen Wei.
\newblock {Towards High Performance Video Object Detection}.
\newblock {\em Proceedings of the IEEE Computer Society Conference on Computer
  Vision and Pattern Recognition}, pages 7210--7218, 2018.

\bibitem{Zhu2017}
Xizhou Zhu, Yuwen Xiong, Jifeng Dai, Lu Yuan, and Yichen Wei.
\newblock {Deep feature flow for video recognition}.
\newblock {\em Proceedings - 30th IEEE Conference on Computer Vision and
  Pattern Recognition, CVPR 2017}, 2017-January:4141--4150, 2017.

\end{thebibliography}
}

\clearpage
\section*{Supplementary Material}
In this document, we include the full object detection evaluation table reporting accuracy and speedup in all configurations (see Figure \ref{tab:acc_speedup_object_detection_full}). 
Furthermore, we explain in more detail how the convolutions are implemented to achieve high efficiency (Section \ref{sec:eff_conv_on_gpus}) and perform ablation studies on threshold tuning (Section \ref{sec:thresholds_ablation}).

\section{Efficient convolution on GPUs}
\label{sec:eff_conv_on_gpus}
Designing an efficient convolutional layer that is on par with today's leading framework, \emph{cuDNN}, requires extensive profiling, analysis and optimizations. 
Minimization of memory transfers and logic operations, as well as optimizing local memory usage (registers and shared memory) are key in achieving a high performance for compute-heavy operations like convolutions.
% Small changes like adding a runtime conditional to be able to skip single inputs that were not updated can increase the average runtime many times.
In this section, we discuss our design decisions and perform ablation studies.

\subsection{Tiling for memory reuse}
Like cuDNN, we perform convolutions in rectangular tiles. 
Each tile consists of a few output pixels (4-48, depending on local memory requirements) and is processed by a single cooperative thread array (CTA).
Neighboring output pixels share most of their input pixels - in the case of a 3x3 filter, 2 neighboring output pixels share 6 out of each one's 9 input pixels.
Larger tiles lead to better memory reuse and can thereby lower the number of global memory transfers significantly.
For example, we use a tile size of 6x6 output pixels for a 3x3 convolutional kernel with a stride of 1.
Compared to processing each pixel individually, this reduces the memory reads to less than a fourth, effectively reading less than 2 inputs per output instead of 9.
Even more important is the reuse of filter parameters which are often larger than input feature maps.
Yet, smaller tile sizes require less registers and allow for more parallelization.
And more importantly, smaller tiles are more likely to be skipped as less inputs can require updates. 
As always, the key is to find the right balance (see Table \ref{tab:tile_size_mode_speed}).

\subsection{Hybrid dense/sparse tile inference}
The convolutional kernel starts with calculating indices, loading the update mask of the input, writing the output mask and, in case any of the inputs was updated, performing the actual convolution.
The convolution is performed in three steps: a) load updated input pixels and store them in CTA shared memory and store zero values for inputs which were not updated, b) load filter values and multiply them with the input stored in shared memory, c) write outputs.
Steps a) and c) use runtime conditionals against input and output boundaries, update mask and dilation.
Step b) uses a highly optimized static code that, independent of the input mask and tile position, always performs all multiply-accumulate operations.
Very sparse tiles with four or less active input pixels, making up about 20\% of non-empty tiles in our tests, are accelerated in a special mode which can process sparse tiles up to 2x faster. 
This mode loads only pixels of the filter weights that are required and iterates over an array of active pixels contrary to iterating over all pixels and checking the update flag.
The performance impact of adding a runtime conditional to allow for per-pixel sparsity (check each pixel if processing is required) and the performance of our hybrid approach compared to a per-tile sparsity (process all pixels or none) approach are reported in Table \ref{tab:tile_size_mode_speed}.

It should be noted that depth-wise convolutions require a special implementation to stay competitive against cuDNN.
In depth-wise convolutions, every output channel only depends on input values of the corresponding channel in the input feature map.
Because of that, input values are used less often and (in our implementation) only by a single thread, removing the necessity of keeping data local to a CTA. 
When loading input data, the update flag of a pixel must always be checked before reading the values because non-updated pixels contain invalid data.
Since the loaded value is only used for up to 9 multiplications in the case of a 3x3 convolution, we fuse loading and multiplication steps described above.
Thus, we decide per pixel if we load the value and perform the multiplication -- resulting in a per-pixel sparse operation compared to per-tile for standard convolutions.
Still, skipping a tile entirely is much faster than processing a single input pixel.

\begin{table}[ptb]
\centering
\setlength{\tabcolsep}{4pt}
% \renewcommand\arraystretch{1.2}
% \resizebox{\linewidth}{!}{
\begin{tabular}{c|c||c|c|c|c}
\hline\thickhline
Tile Mode & Tile Size & s=0\% & s=50\% & s=90\% & s=99\% \\
\hline
\multirow{2}{*}{p.t. Sparse}  & 6x6   & 17.0 & 16.9 & 16.5 & 7.7 \\
                        & 5x5   & 23.7 & 23.6 & 22.9 & 8.5 \\
                        % & 4x4   & 18.7 & 18.6 & 17.9 & 5.8  \\
% \cline{1-6}
\hline
\multirow{2}{*}{Hybrid} & 6x6   & 17.0 & 16.9 & 15.8 & 5.0  \\
                        & 5x5   & 23.5 & 23.2 & 18.3 & 4.5  \\
                        % & 4x4   & 18.7 & 18.7 & 15.6 & 4.4  \\
% \cline{2-6}
\hline
\multirow{2}{*}{p.p. Sparse} 
                        & 6x6   & 60.4 & 51.9 & 44.3 & 20.4 \\
                        & 5x5   & 49.6 & 43.6 & 38.0 & 14.1 \\
                        % & 4x4   & 45.4 & 38.6 & 32.5 & 10.2 \\
\hline\thickhline
\end{tabular}
\caption{
GTX 1050 runtime comparison between different tile sizes and sparsity levels ($s$) reported in milliseconds for a 3x3 convolution with 128 input and output channels and a 256x256 pixels input.
Since sparsity is generated uniformly, even with a 90\% sparse input is very likely have at least one updated pixel per tile.
\emph{Per-tile sparse} mode always processes either all pixels of a tile or none of them. 
\emph{Per-pixel sparse} mode decides per input pixel whether it needs to be processed or can be skipped.
\emph{Hybrid} mode uses dense mode when more than four input pixels are updated and uses a special implementation for very sparse tiles.
Larger tile sizes cannot be used due to local memory resource limitations.
}
% }
\label{tab:tile_size_mode_speed}
\end{table}

\subsection{Memory layout for bandwidth reductions}
Using a per-pixel update mask, we can reduce the memory bandwidth during inference greatly in many ways.
Compared to previous work that had to compare the inputs of each layer against previous values, we only load values that were marked as updated.
Furthermore, using the update mask, we do not have to set unchanged values in the feature map to zero.
This allows us to use uninitialized memory for the greater part of the feature maps - only writing valid values for updated pixels.
Compared to setting all unchanged output features to zero, this improved the performance of convolutions in HRNet by up to 68\%.

PyTorch's default memory layout is $NCHW$, storing one image per pixel channel.
Accessing pixels individually, however, is very inefficient with this layout, as the minimum memory access size -- a cache line -- would always load multiple neighboring pixels per instruction. 
For efficient use of per-pixel sparsity, we store feature maps in $NHWC$ format.
This way, all channels of a pixel are stored coalesced and can be accessed efficiently.

\subsection{Floating point number inaccuracies}
While delta updates can in theory be applied indefinitely, floating point number operations result in slightly different outcomes when taking large accumulated values or small deltas as input.
With HRNet and Human3.6M, we did not experience any problems even with sequences thousands of frames long.
However, we did notice small errors accumulating over time in EfficientDet, even with dense inference, \ie using negative thresholds.
We recommend to reset the buffers every few hundred frames, or when the network input switches between different videos, to flush all accumulated errors due to floating point inaccuracies.

\section{Threshold tuning ablation study}
\label{sec:thresholds_ablation}
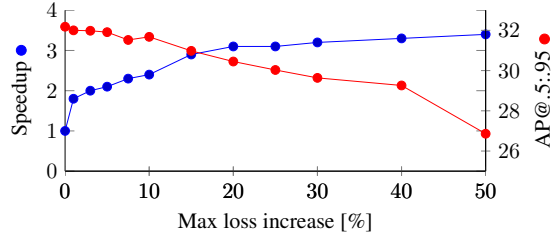
\begin{figure}
\centering
\resizebox{0.9\linewidth}{!}{
\begin{tikzpicture}
\begin{axis}[
ylabel={Speedup \tikz\draw[blue,fill=blue] (0,0) circle (.5ex);},
axis x line=bottom,
axis line style={-},
ymin=0,
ymax=4,
width=\linewidth,
height=0.5\linewidth,
]
\addplot
coordinates {
(0,     1.0)
(1,     1.8)
(3,     2.0)
(5,     2.1)
(7.5,   2.3)
(10,    2.4)
(15,    2.9)
(20,    3.1)
(25,    3.1)
(30,    3.2)
(40,    3.3)
(50,    3.4)
};
\end{axis}
\begin{axis}[
axis y line=right,
axis x line=bottom,
axis line style={-},
ylabel={AP@.5:.95 \tikz\draw[red,fill=red] (0,0) circle (.5ex);},
xlabel={Max loss increase [\%]},
ymin=25,
ymax=33,
width=\linewidth,
height=0.5\linewidth,
]
\addplot[
color=red,
mark=*,
mark options={solid},
]
coordinates {
(0,     32.18)
(1,     32.00)
% (2,     31.95)
(3,     31.98)
(5,     31.91)
(7.5,   31.522)
(10,    31.677)
(15,    30.98)
(20,    30.45)
(25,    30.03)
(30,    29.637)
(40,    29.26)
(50,    26.86)
% (0,     63.9)
% (1,     63.6)
% (3,     63.4)
% (5,     63.9)
% (7.5,   62.9)
% (10,    63.1)
% (15,    62.1)
% (20,    62.0)
% (25,    61.6)
% (30,    61.3)
% (50,    58.3)
};
\end{axis}
\end{tikzpicture}
} % resizebox
\caption{Accuracy and speedup achieved with EfficientDet-d1 on the MOT16 dataset with thresholds tuned on different maximum loss increase targets. Even when using a 30\% total loss increase target, the accuracy stays close to dense inference. With a smaller step size for threshold tuning, the accuracy could be kept closer to the maximum loss increase target.
% \todo{move to supplementary material} \todo{try scatter plot here?}
% \todo{drop max loss increase axis and only use speedup vs AP as scatters}
}
\label{fig:tuning_thresholds}
\end{figure}

In our evaluations, we used a maximum loss increase target of 3\% for the task of threshold tuning.
To ensure that we do not exceed this limit, every truncation threshold is only allowed to increase the loss by a maximum of $\frac{3\%}{\#_{layers}}$.
Due to a predefined step size in the threshold tuning process, however, the actual loss increase is typically much lower.
We evaluated different maximum loss increase targets to compare the resulting accuracy and speedup (see Figure \ref{fig:tuning_thresholds}).
To better show the impact of different parameters, we do not use a high threshold and update mask dilation on the first layer as this already increases sparsity greatly, and thereby reduces the impact of the chosen parameters.
Instead, we set the first threshold to a low value of 0.15 manually.

\subsection{Training truncation thresholds}
We experimented with training truncation thresholds in parallel instead of auto-tuning them one-by-one. 
Using the update mask density as an additional loss, the trade-off between accuracy and pixel update density could be optimized more accurately and the accuracy reduction per layer could be better balanced. 
\Eg early layers of the CNNs can be allowed a larger accuracy decrease, because they have a larger impact on the overall update density. 

Instead of boolean update masks, we used soft truncation with a sigmoid activation function to allow for gradients to propagate better.
With the same intention, we switched to using norm truncation instead of maximum truncation, \ie we compare the $L^2$ norm of a pixel's \emph{deltas} against a threshold.
This way, all values contribute to the update mask and the training process is more stable than with maximum truncation.
While we managed to train feasible thresholds with a pre-defined accuracy vs. sparsity trade-off, the resulting thresholds did not outperform tuned thresholds.
Furthermore, threshold training takes many times longer to process and hyper parameters are more difficult to tune than the single parameter used for auto tuning.

\begin{table*}[ptb]
\centering
\setlength{\tabcolsep}{4pt}
\renewcommand\arraystretch{1.2}
\resizebox{\linewidth}{!}{
\begin{tabular}{cc||c|c|c||cc|cc|cc|cc|cc}
\hline\thickhline
\multirow{2}{*}{Dataset} & \multirow{2}{*}{Backend} & \multirow{2}{*}{AP@0.5} & \multirow{2}{*}{AP@.5:.95} & \multirow{2}{*}{GFLOPs} & \multicolumn{2}{c|}{Jetson Nano b=1} & \multicolumn{2}{c|}{GTX 1050 b=1} & \multicolumn{2}{c|}{GTX 1050 b=8/3} & \multicolumn{2}{c|}{RTX 3090 b=1} & \multicolumn{2}{c}{RTX 3090 b=48/24} \\
\cline{6-15}
& & & & & FPS & speedup & FPS & speedup & FPS & speedup & FPS & speedup & FPS & speedup \\
\hline
\multirow{3}{*}{MOT16-d0} 
& cuDNN         & \multirow{2}{*}{55.6\%}   & \multirow{2}{*}{27.8\%} & \multirow{2}{*}{5.0}
                                                 & 1.4  & 1.0 & 26.0 & 1.0 & 26.4 & 1.0 & 29.3   & 1.0 & 368    & 1.0 \\
& ours dense            &           &           && 4.0  & 2.9 & 29.3 & 1.1 & 28.6 & 1.1 & 57.2   & 2.0 & 369    & 1.0 \\
& ours $\epsilon=\infty$& 17.9\%    & 6.6\%     & -     & 8.8   & 6.3 & 61.0 & 2.4 & 340  & 12.9& 57.0   & 2.0 & 2389   & 6.5 \\
& CBInfer               & 54.3\%    & 27.0\%    & 1.7   & 2.0   & 1.4 & 9.5  & 0.4 & 24.3 & 0.9 & 15.6   & 0.5 & 273    & 0.7 \\
% & ours sparse 1         & 56.0\%    & 27.7\%    & -     & -   & 38.1 & 1.5 & 56.9 & 2.2 & 57.9   & 2.0 & 692   & 1.9 \\
% & ours sparse 2         & 55.2\%    & 27.0\%    & -     & -   & 47.0 & 1.8 & 75.3 & 2.9 & 58.0   & 2.0 & 957   & 2.6 \\
% & ours sparse norm 1    & 55.7\%    & 27.4\%    & -     & -   & -    & -   & -    & -   & -      & -   & 777   & 2.1 \\
% & ours sparse norm 2    & 55.5\%    & 27.6\%    & -     & -   & -    & -   & -    & -   & -      & -   & 664   & 1.8 \\
& ours sparse           & 55.8\%    & 27.8\%    & 2.8   & 7.9   & 5.6 & 48.2 & 1.9 & 76.6 & 2.9 & 57.2    & 2.0 & 896   & 2.4 \\
% & ours sparse 3\% *     & 55.8\%    & 27.8\%    & -     & -   & 48.2 & 1.9 & 76.6 & 2.9 & -      & -   & 896   & 2.4 \\
% & ours sparse 5\% *     & 55.1\%    & 27.3\%    & -     & -   & -    & -   & -    & -   & -      & -   & 927   & 2.5 \\
% & ours sparse 10\% *    & 54.2\%    & 26.6\%    & -     & -   & -    & -   & -    & -   & -      & -   & 1080  & 2.9 \\
\hline
\multirow{3}{*}{MOT16-d1} 
& cuDNN         & \multirow{2}{*}{63.9\%}    & \multirow{2}{*}{32.2\%} & \multirow{2}{*}{12.2}
                                                        & 0.7   & 1.0 & 10.5 & 1.0 & 11.3   & 1.0 & 23.3 & 1.0  & 169 & 1.0 \\
& ours dense            &           &           &       & 1.8   & 2.6 & 11.6 & 1.1 & 11.8   & 1.0 & 45.0 & 1.9  & 167 & 1.0 \\
& ours $\epsilon=\infty$& 21.7\%    & 7.2\%     & - & 6.6   & 9.4 & 51.1 & 4.9 & 151.6  & 13.4& 44.0 & 1.9  & 1032& 6.1 \\
& CBInfer               & 63.5\%    & 31.9\%    & 4.6   & 1.3   & 1.9 & 6.3  & 0.6 & 9.4    & 0.8 & 11.4 & 0.5  & 106 & 0.6 \\
% & CBInfer 2             & 63.1\%    & 32.3\%    & -     & -   & 6.3  & 0.6 & 9.5    & 0.8 & -    & -    & 104 & 0.6 \\
& ours sparse           & 64.0\%    & 32.0\%    & 6.8   & 3.9   & 5.6 & 27.4 & 2.6 & 30.9   & 2.7 & 45.2 & 1.9  & 389 & 2.3 \\
\hline
% \multirow{3}{*}{MOT16-d4} 
% & cuDNN         & \multirow{2}{*}{67.8\%}   & \multirow{2}{*}{39.7\%}       
%                                         & -     & 1.0 & -   & 1.0 & -   & 1.0 & -   & 1.0 & 28.3   & 1.0 \\
% & ours dense    &           &           & -     & -   & -   & -   & -   & -   & -   & -   & 29.3   & 1.0 \\
% & ours sparse   & 69.2\%    & 39.7\%    & -     & -   & -   & -   & -   & -   & -   & -   & 56.4   & 2.0 \\
% \hline
\multirow{3}{*}{WildTrack-d0} 
& cuDNN         & \multirow{2}{*}{67.1\%}   & \multirow{2}{*}{35.4\%} & \multirow{2}{*}{5.0}
                                            & 1.4   & 1.0 & 26.0& 1.0 & 26.4    & 1.0   & 29.3  & 1.0   & 383   & 1.0  \\
& ours dense        &           &           && 4.0  & 2.9 & 29.3& 1.1 & 28.6    & 1.1   & 57.2  & 2.0   & 384   & 1.0  \\
& ours $\epsilon=\infty$& 29.5\%& 14.3\%    & -     & 8.8  & 6.3 & 61.0& 2.4 & 340     & 12.9  & 57.0  & 2.0   & 2389  & 6.5 \\
& CBInfer           & 65.4\%    & 33.9\%    & 1.1 & 2.2  & 1.6 & 10.4& 0.4 & 25.0    & 0.9   & 16.8  & 0.6   & 274   & 0.7  \\
% & ours sparse       & 66.5\%    & 34.5\%    & -     & -   & -   & -   & 82.6    & 3.1   & -   & -   & 1017  & 2.6  \\
& ours sparse       & 66.4\%    & 34.7\%    & 2.3   & 8.1   & 5.8 & 56.1& 2.2 & 100     & 3.8   & 57.2  & 2.0   & 973   & 2.5  \\
\hline
\multirow{3}{*}{WildTrack-d1} 
& cuDNN         & \multirow{2}{*}{72.1\%}   & \multirow{2}{*}{41.6\%} & \multirow{2}{*}{12.2}
                                            & 0.7   & 1.0   & 10.5    & 1.0 & 11.3  & 1.0   & 23.3  & 1.0     & 169   & 1.0  \\
& ours dense        &           &           && 1.8   & 2.6   & 11.6    & 1.1 & 11.8  & 1.0   & 45.0  & 1.9   & 167   & 1.0  \\
& ours $\epsilon=\infty$& 28.3\%& 15.0\%    & -     & 6.6   & 9.4   & 51.1    & 4.9 & 151   & 13.4  & 45.9  & 2.0   & 1032  & 6.1 \\
% & CBInfer           & 71.8\%         & 41.2\%    & -     & -   & 10.2    & 1.0   &        & -     & -     & -     & -     & -  \\
& CBInfer           & 71.5\%    & 40.5\%    & 2.4   & 1.4   & 2.0   & 6.8     & 0.6 & 10.6  & 0.9   & 13.1  & 0.6   & 120   & 0.7 \\
% & ours sparse       & 71.6\%    & 40.9\%    & -     & -   & 24.9    & 2.4   & 26.3  & 2.3  & 46.0  & 2.0 & 360  & 2.1  \\
& ours sparse       & 71.6\%    & 40.8\%    & 5.7   & 4.8   & 6.9   & 32.1    & 3.1 & 38.6  & 3.4   & 44.8  & 1.9   & 415   & 2.5  \\
% \hline
% \multirow{3}{*}{WildTrack-d4} 
% & cuDNN         & \multirow{2}{*}{72.4\%}   & \multirow{2}{*}{52.3\%}       
%                                         & -     & 1.0 & -   & 1.0 & -   & 1.0 & -   & 1.0 & 28.3   & 1.0  \\
% & ours dense    &           &           & -     & -   & -   & -   & -   & -   & -   & -   & 29.3   & 1.0  \\
% % & ours sparse   & 71.8\% \todo{update}    & 50.0\% \todo{update}    & -     & -   & -   & -   & -   & -   & -   & -   & -       & -  \\
% & ours sparse   & 72.9\%    & 49.7\%    & -     & -   & -   & -   & -   & -   & -   & -   & -       & -  \\
\hline\thickhline
\end{tabular}
}
\caption{Speed and accuracy comparisons of different CNN backends used for the task of object detection with batch size $b$.
MOT16 is evaluated only on fixed camera videos. 
Our evaluations show that \ours~achieves higher frame rate and accuracy using the 12.2 GFLOPs \emph{d1} configuration of EfficientDet than cuDNN achieves with the 5.0 GFLOPs {d0} configuration.
% \todo{I split the train set into train and test set with a 80/20 ratio, because the labels for train set are called GT and the labels for test set are called DET. Find out if the actual test set is also correctly labeled!}
% \chengcheng{Highlight the best results?}
% \todo{only report FPS for cudnn - speedup for rest}
% \todo{shorter table titles}
% \todo{for d1, we only need ours sparse line}
% \todo{put full version in supplementary material}
% \todo{discuss in meeting if we want to keep e=infinity?}
% https://tex.stackexchange.com/questions/32683/rotated-column-titles-in-tabular?fbclid=IwAR21O9Qs2h9TMOO2Wu283IUnlovGZKhZfJ_fG4guArWQ3NDYSdD8Srfgcmw
}
\label{tab:acc_speedup_object_detection_full}
\end{table*}

\end{document}